\documentclass[conference]{IEEEtran}

\usepackage{graphicx}
\usepackage{booktabs}
\usepackage{float}
\usepackage{subcaption}
\usepackage{pifont}
\usepackage{placeins}
\usepackage{amsmath}
\usepackage{amssymb}
\usepackage{hyperref}
\usepackage{cite}

\begin{document}

\title{Virtual Try-On for Cultural Clothing: A Benchmarking Study}

\author{
\IEEEauthorblockN{
Muhammad Tausif Ul Islam, Shahir Awlad, Sameen Yeaser Adib \\
\vspace{0.2cm} 
Md. Atiqur Rahman, Sabbir Ahmed, Md Hasanul Kabir
}
}
\maketitle

\begin{abstract}
Although existing virtual try-on systems have made significant progress with the advent of diffusion models, the current benchmarks of these models are based on datasets that are dominant in western-style clothing and female models, limiting their ability to generalize culturally diverse clothing styles. In this work, we introduce BD-VITON, a virtual try-on dataset focused on Bangladeshi garments, including saree, panjabi and salwar kameez, covering both male and female categories as well. These garments present unique structural challenges such as complex draping, asymmetric layering, and high deformation complexities which are underrepresented in the original VITON dataset. To establish strong baselines, we retrain and evaluate try-on models, namely StableViton, HR-VITON, and VITON-HD on our dataset. Our experiments demonstrate consistent improvements in terms of both quantitative and qualitative analysis, compared to zero-shot inference.
\end{abstract}


\section{Introduction}
\label{sec:intro}

Advancements in the VITON technology has opened vast possibilities across fashion and e-commerce industries by enabling customers to visualize garments on themselves without having to physically try them on in fitting rooms. Over the past several years, a succession of increasingly sophisticated models - from VITON-HD~\cite{choi2021vitonhd} to HR-VITON~\cite{lee2022hrviton} and StableVITON~\cite{kim2023stableviton} - have demonstrated remarkable photorealistic quality in compositing garments onto person images, driven by recent progress in generative modeling and pose estimation. Such systems also enhance online shopping experiences and hold considerable promise for markets where physical retail infrastructure is limited.

Nevertheless there is a significant limitation that continues to persist: all existing try-on research and more importantly the dataset they rely on are constructed exclusively around Western fashion conventions. The dominant benchmarks in the field - VITON~\cite{han2018viton} and its derivatives - extract garment and person images from online retailers, reflecting clothing styles and garment structures of the Western apparel. This narrow diversity in the dataset results in trained models which systematically fail when confronted with different draping behaviors and wearing conventions that fall outside their recognition. Consequently, for the 2.4 billion people in South Asia who regularly wear garments tied to their own traditions, the existing try-on systems render ineffective.

To address this gap, we introduce BD-VITON, a virtual try-on dataset focused on Bangladeshi traditional attire. Our dataset consists of paired garment--person image pairs covering sarees, panjabis, and kameez on both male and female subjects, collected and curated for training and evaluating existing VITON architectures.

The central contribution of this work is to address a foundational argument; culturally specific, in-distribution training data can significantly improve the performance, when the model architecture remains unaltered. We demonstrate this by fine tuning three state-of-the-art VITON models - VITON-HD~\cite{choi2021vitonhd}, HR-VITON~\cite{lee2022hrviton} and StableVITON~\cite{kim2023stableviton} - on our dataset and evaluating them against zero-shot inferences. Across all the architectures, meaningful improvements hinted that there is a distributional mismatch and that the dataset construction is effectively overcoming this gap.

\section{Related Works}

The past few years have witnessed rapid growth in the try-on models. Primitive models such as CP-VTON~\cite{wang2018cpviton}, VITON-HD~\cite{choi2021vitonhd} heavily relied on Thin Plate Spline transformation (TPS), addressing preservation of cloth textures when wearers had more complex poses. Their successor, HR-VITON~\cite{lee2022hrviton} used target human parsing maps alongside TPS to structurally guide the warping of the cloth, rendering it to be more region aware instead of being globally warped around the entire person. As of 2024, StableVITON~\cite{kim2023stableviton} abandoned the TPS transformation and adhered to latent feature representations along with cross-attention diffusion model to guide the architecture so as to correctly warp the cloth onto the target person. Regardless of how well the architecture of these try-on models improved over the years, the fact that they were mainly trained on a dataset that is tunnel visioned on upper-body western style clothing of predominant female models makes it difficult to evaluate the full potential of these try-on models. BD-VITON is designed to address this by introducing Bangladeshi styled attires which introduce more complex garment structure with increased folds and drapes so that existing try-on models can also be utilized effectively for a broader set of real-world clothing scenarios.

One of the more recent datasets, DressCode~\cite{morelli2022dresscode}, successfully addresses several issues in the original VITON dataset. The previous VITON dataset included only upper-body clothing, completely ignoring lower-body clothing. DressCode~\cite{morelli2022dresscode} addresses this by including lower-body clothing despite the limitation of samples in this field. The dataset also introduces a larger number of samples, 3x to be precise, relative to the VITON dataset. Additionally, the VITON dataset is predominant with female model images, leading to poor evaluation for male clothing, DressCode~\cite{morelli2022dresscode} has also successfully addressed this by introducing both male and female clothing of adequate proportions.

Deep Fashion3D~\cite{xu2020deepfashion3d} introduces a unique dataset by introducing 3D models for a wide range of applications including virtual try-on models. They include a total of 2078 3D models spread across 10 categories with 563 garment instances, covering a wide range of both top and bottom clothing. Such real life 3D scans of garments and models allow for quality virtual try-on applications. Deep Fashion3D~\cite{xu2020deepfashion3d} also introduces an abundance of rich annotations, 3D feature lines, 3D body poses with images viewed from multiple angles. This gives access to dynamic pose variability in try-on experiences.

Nevertheless, a common loophole continues to exist - lack of clothing that have complex garment structures, potentially imposing a greater challenge to try-on models. Since the recently researched datasets aforementioned also base their research on Western apparel, it is difficult to conclude whether the try-on models would actually perform optimally if a more complex clothing is presented. Such types of clothing are mainly available in Asian regions. To address this gap in the field, we introduce BD-VITON, which introduces more complex clothing, derived from Bangladeshi culture.

\section{Dataset Construction}

\begin{figure*}[t]
\centering
\begin{subfigure}[b]{0.321\textwidth}\centering
  \includegraphics[width=\linewidth,height=6cm,keepaspectratio]{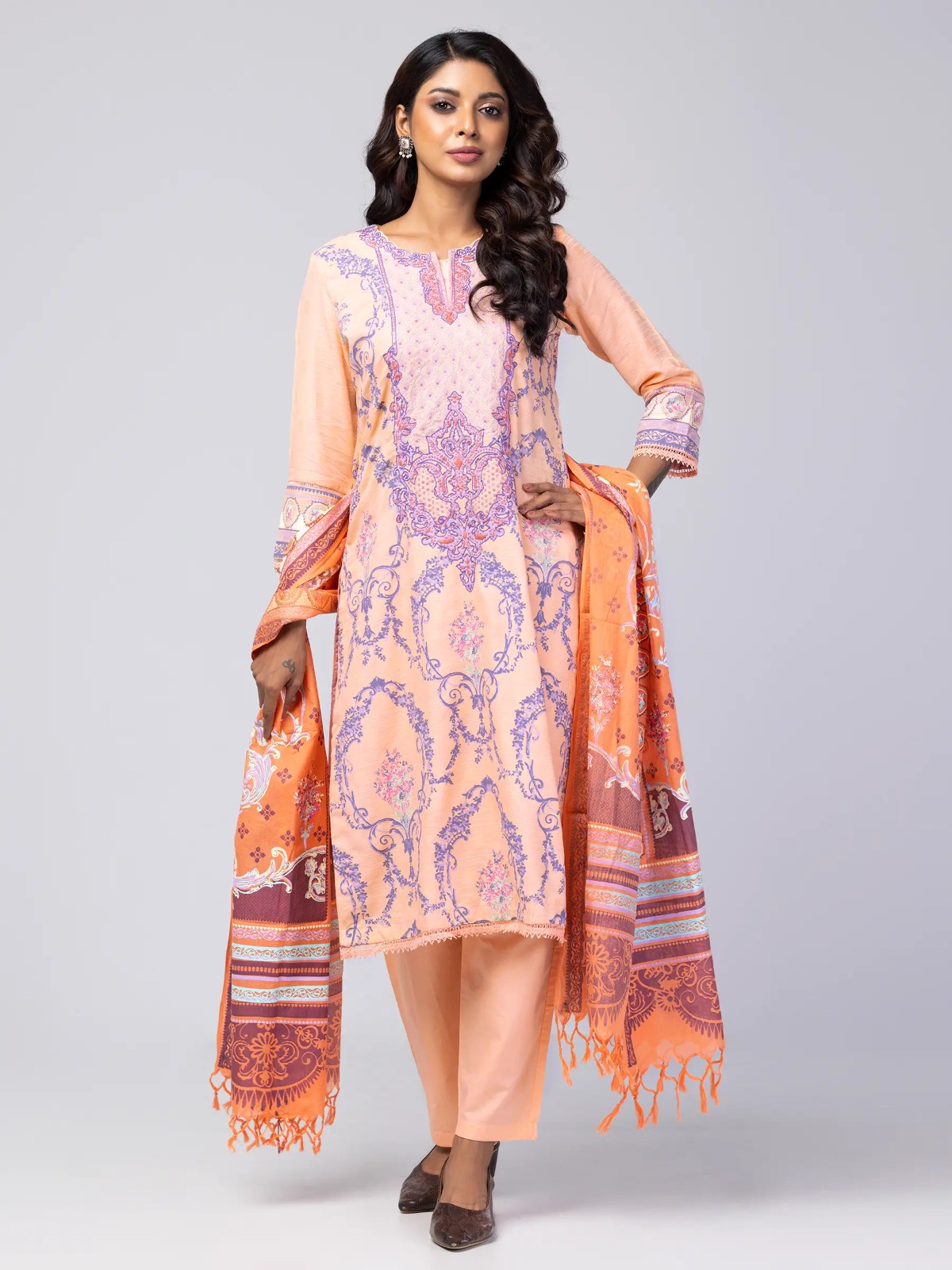}
\end{subfigure}\hfill
\begin{subfigure}[b]{0.285\textwidth}\centering
  \includegraphics[width=\linewidth,height=6cm,keepaspectratio]{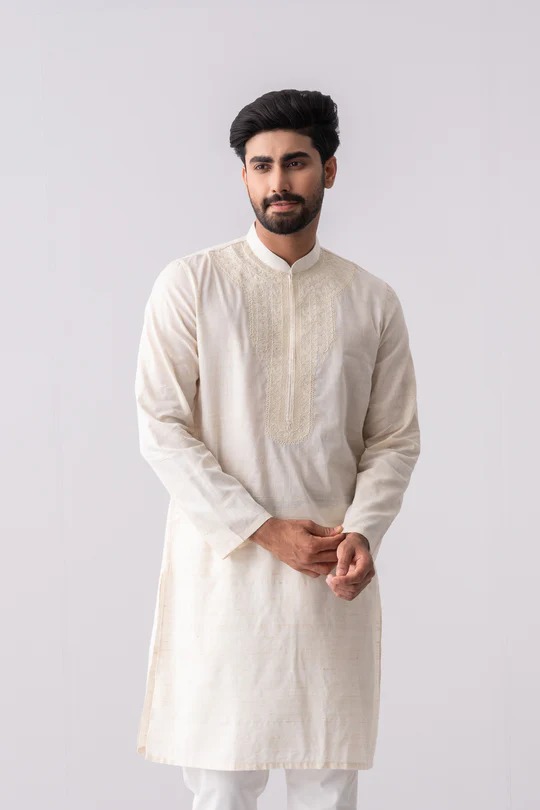}
\end{subfigure}\hfill
\begin{subfigure}[b]{0.356\textwidth}\centering
  \includegraphics[width=\linewidth,height=6cm,keepaspectratio]{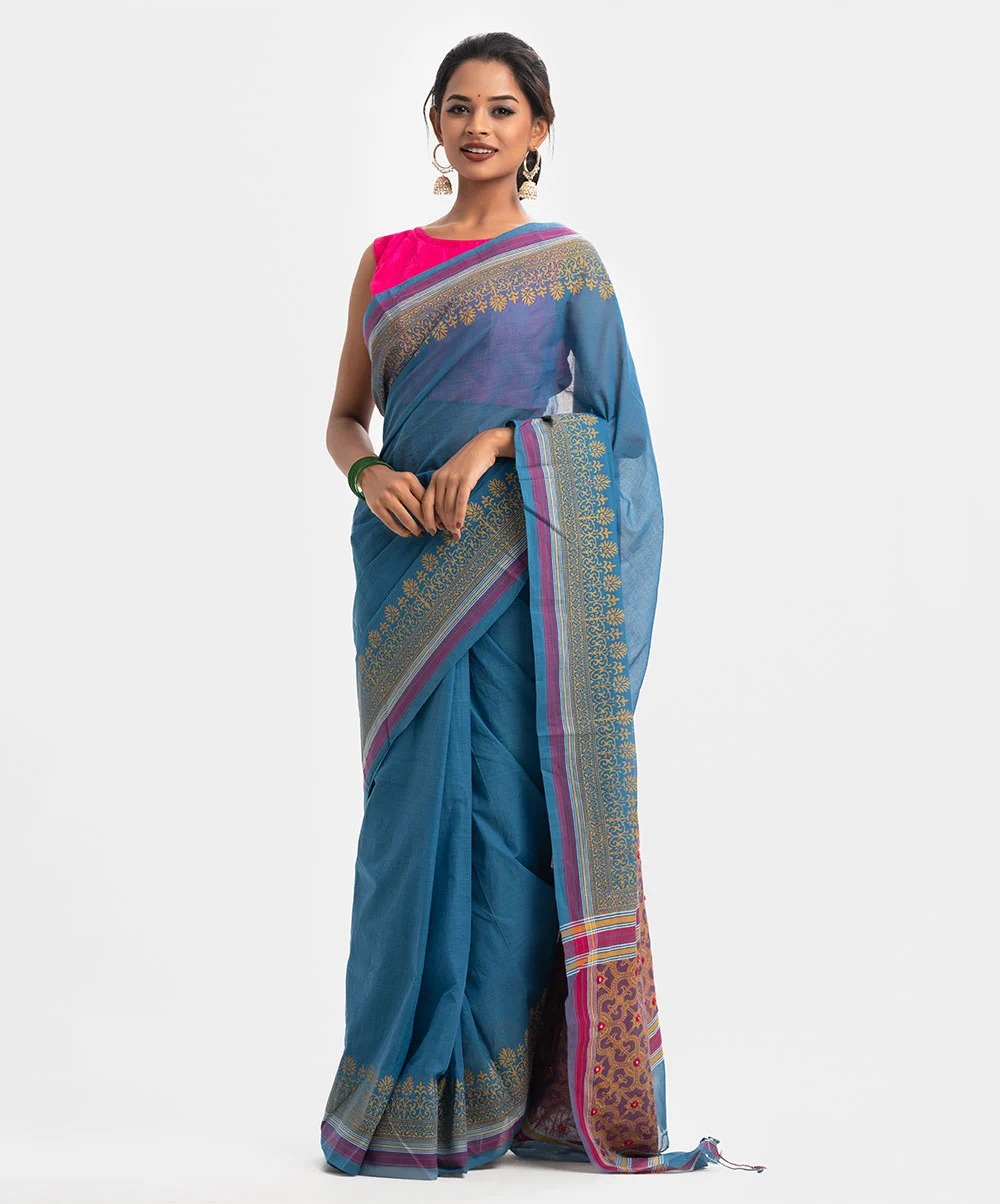}
\end{subfigure}
\caption{Samples of BD-VITON, Kameez, Panjabi and Saree from left to right}
\label{fig:dataset_composition_showcase}
\end{figure*}

\subsection{Dataset Motive}
Bangladesh presents such an underrepresented case. Traditional Bangladeshi attire encompasses a wide and geometrically complex set of garments - the saree, an unstitched continuous drape that wraps around the wearer's body through several intricate folds; the panjabi, a collarless tunic predominantly worn by men; and the kameez, a long tailored garment worn mainly by women, often paired with a separate overlaying garment, orna. The aforementioned types of clothing exhibit substantial variability in topology, body coverage and wearing semantics.

\subsection{Data Collection Process}
We first collect only person images from publicly available Bangladeshi online retail platforms that specialize in the targeted attire. All images were sourced from publicly accessible product listings. To ensure diversity and quality, we filtered images to include uniform white backgrounds, varying model poses and visually distinct garments. By collecting panjabi samples, we ensure that male models are also covered in our dataset. For training purposes, we decided to split the dataset in an 80/20 split, with 810 samples preserved for training and 203 samples for testing purposes.

\begin{table}[H]
\centering
\caption{BD-VITON dataset composition by gender and attire type.}
\setlength{\tabcolsep}{12pt}
\begin{tabular}{lcc}
\toprule
\textbf{Clothing Type} & \textbf{Male Samples} & \textbf{Female Samples} \\
\midrule
Panjabi                & 340                   & -                     \\
Kameez                 & -                     & 341                   \\
Saree                  & -                     & 332                   \\
\midrule
\textbf{Total}         & 340                   & 673                   \\
\bottomrule
\end{tabular}
\label{tab:bdviton_distribution}
\end{table}

The gender imbalance is due to relative scarcity of traditional male Bangladeshi attire in commercial platforms and the fact that majority attire of males in general is Western styled with only panjabi being a proper traditional wear. In contrast, female attire is more widely available and diverse, resulting in a larger sample count.

\begin{figure}[H]
    \centering
        \includegraphics[width=\columnwidth]{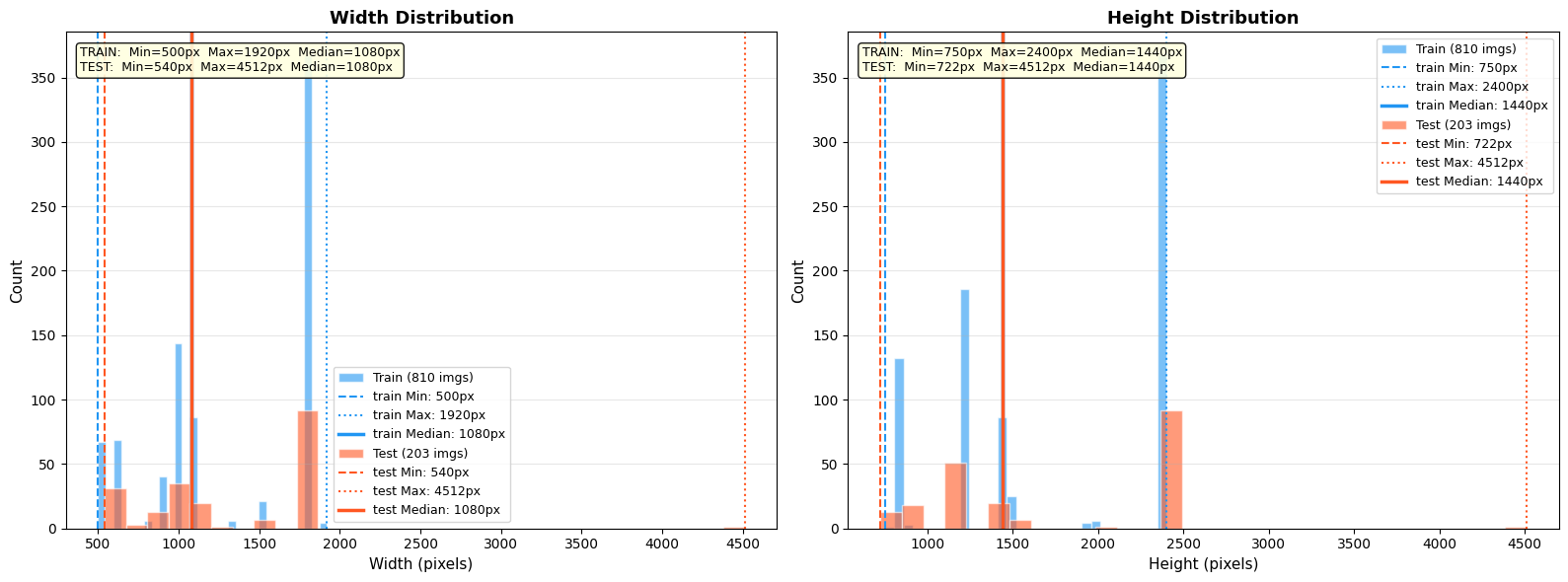}
        \caption{Resolution distribution of BD-VITON (Train+Test).}
        \label{fig:reslabel}
\end{figure}

The samples in BD-VITON exhibit a wide range of resolutions as depicted in Figure~\ref{fig:reslabel}, with image widths from 500px to 4512px and heights ranging from 722px to 4512px. With a median resolution of 1,080$\times$1,440, we ensure BD-VITON is well-suited for high resolution try-on evaluation.

As shown in Table~\ref{tab:dataset_comparison}, existing try-on datasets are only restricted to Western apparels and female subjects. The original VITON dataset is limited to only upper body garments, with relatively lower resolutions, working exclusively on female subjects. DressCode~\cite{morelli2022dresscode}, one of the more recent works on datasets, represent gender coverage with full-body garment support, yet is only based on Western fashion domain, once again leaving concerns to how try-on models would actually perform on clothing of much more complex structures. BD-VITON is distinguished in the sense that it introduces a new culture of clothing style while also maintaining the shortcomings that DressCode~\cite{morelli2022dresscode} has addressed.

\begin{table}[h]
\centering
\caption{Comparison of BD-VITON with existing virtual try-on datasets.}
\resizebox{\columnwidth}{!}{%
\begin{tabular}{lcccccc}
\toprule
\textbf{Dataset} & \textbf{Pairs} & \textbf{Median Res.} & \textbf{Gender} & \textbf{Body Coverage} & \textbf{Garment Types} & \textbf{Cultural Scope} \\
\midrule
VITON            & 16,253  & 256$\times$192   & F only & Upper-body & Tops, shirts         & Western \\
VITON-HD         & 13,679  & 1024$\times$768  & F only & Upper-body & Tops, shirts         & Western \\
DressCode        & \textbf{53,792}  & 1024$\times$768  & \textbf{M+F}    & \textbf{Full-body}  & Tops, bottoms, t shirts, skirts & Western \\
\textbf{BD-VITON (Ours)} & 1,013 & \textbf{1080$\times$1440} & \textbf{M+F} & \textbf{Full-body} & \textbf{Saree, Panjabi, Kameez} & \textbf{Bangladeshi} \\
\bottomrule
\end{tabular}}
\label{tab:dataset_comparison}
\end{table}

\subsection{Annotation Pipeline}
The composition of BD-VITON follows the structure of the VITON dataset that is used by VITON-HD~\cite{choi2021vitonhd} and its successors as well. Figure~\ref{fig:dataset_composition} gives a sample showcase of masks generated based on one of the samples collected for BD-VITON. For the training and evaluation of HR-VITON~\cite{lee2022hrviton}, we constructed a fully automated annotation pipeline that generates structural and semantic signals from raw person which done with the help of pretrained models.

\subsubsection{Semantic Human Parsing}
\begin{figure*}[t]
    \centering
    \begin{subfigure}{0.19\textwidth}
        \includegraphics[width=\linewidth]{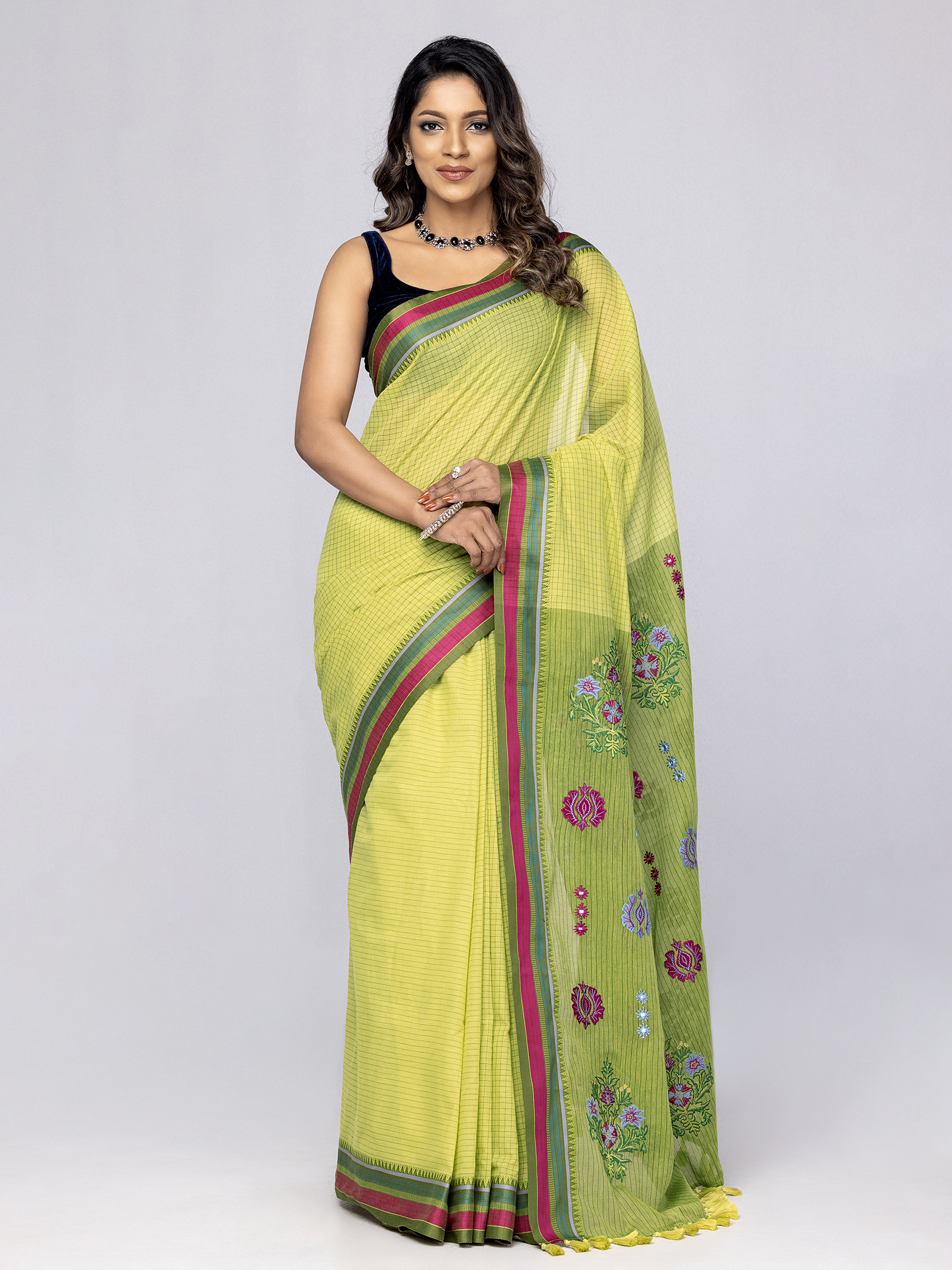}
    \end{subfigure}
    \hfill
    \begin{subfigure}{0.19\textwidth}
        \includegraphics[width=\linewidth]{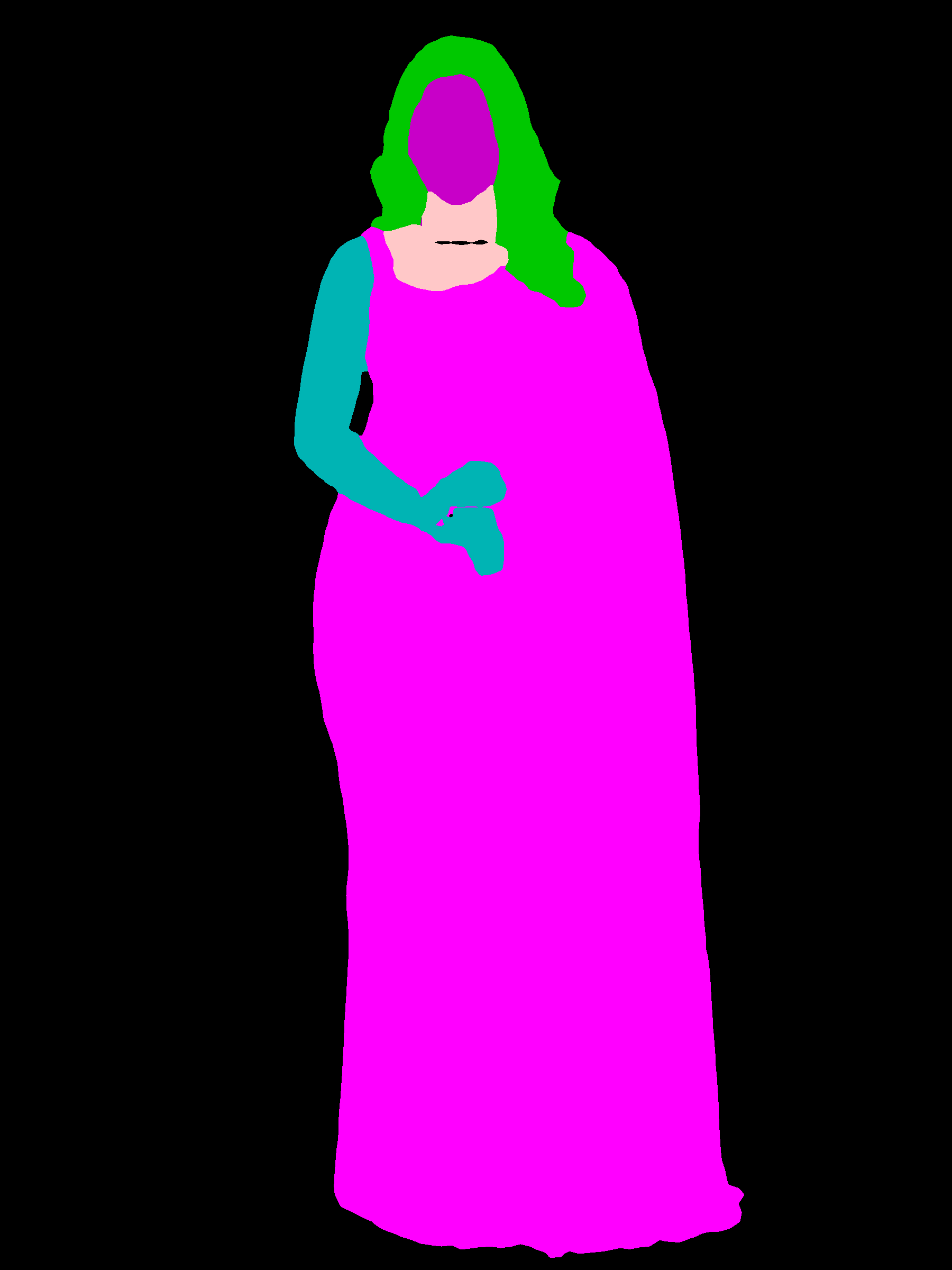}
    \end{subfigure}
    \hfill
    \begin{subfigure}{0.19\textwidth}
        \includegraphics[width=\linewidth]{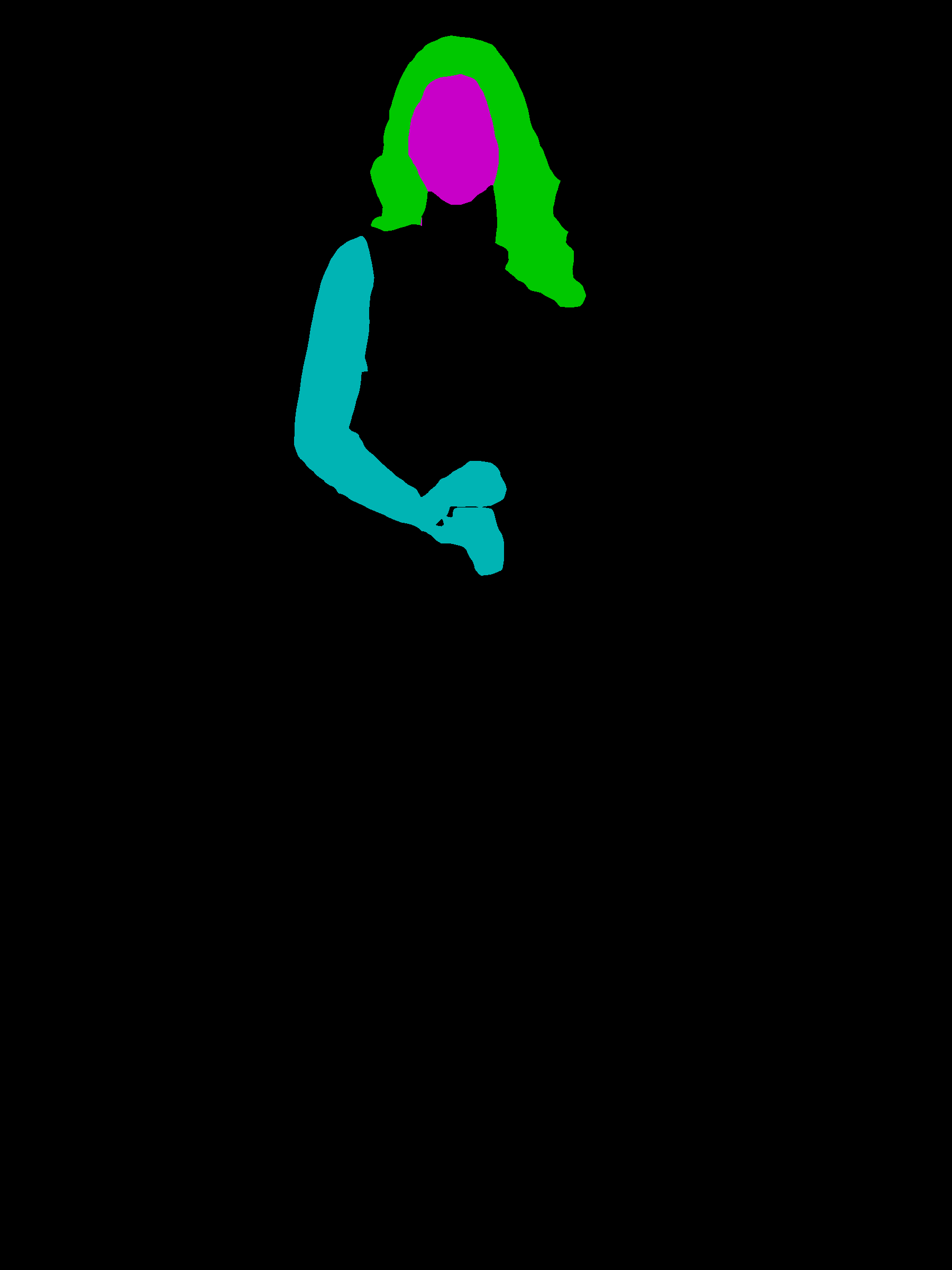}
    \end{subfigure}
    \hfill
    \begin{subfigure}{0.19\textwidth}
        \includegraphics[width=\linewidth]{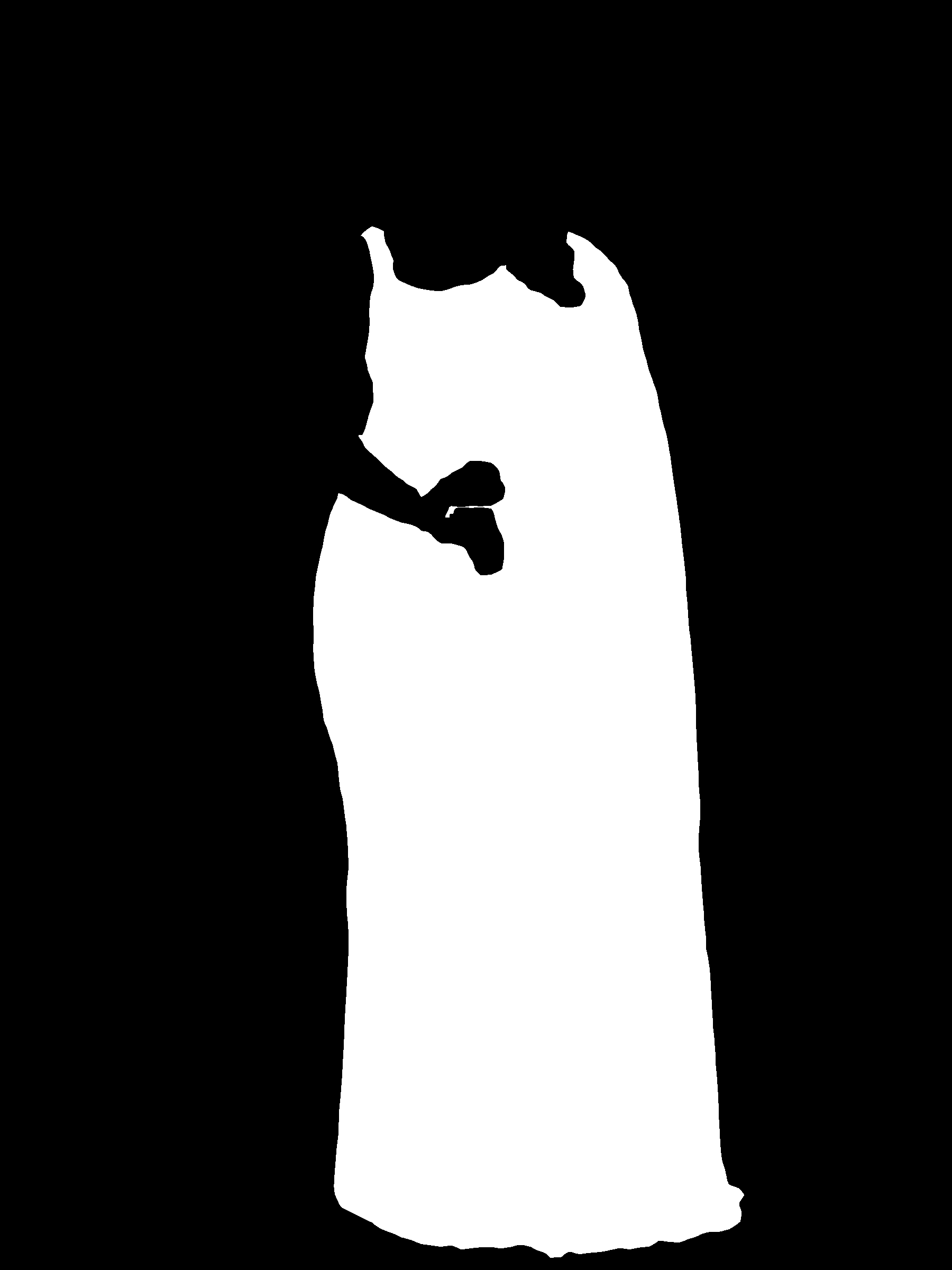}
    \end{subfigure}
    \hfill
    \begin{subfigure}{0.19\textwidth}
        \includegraphics[width=\linewidth]{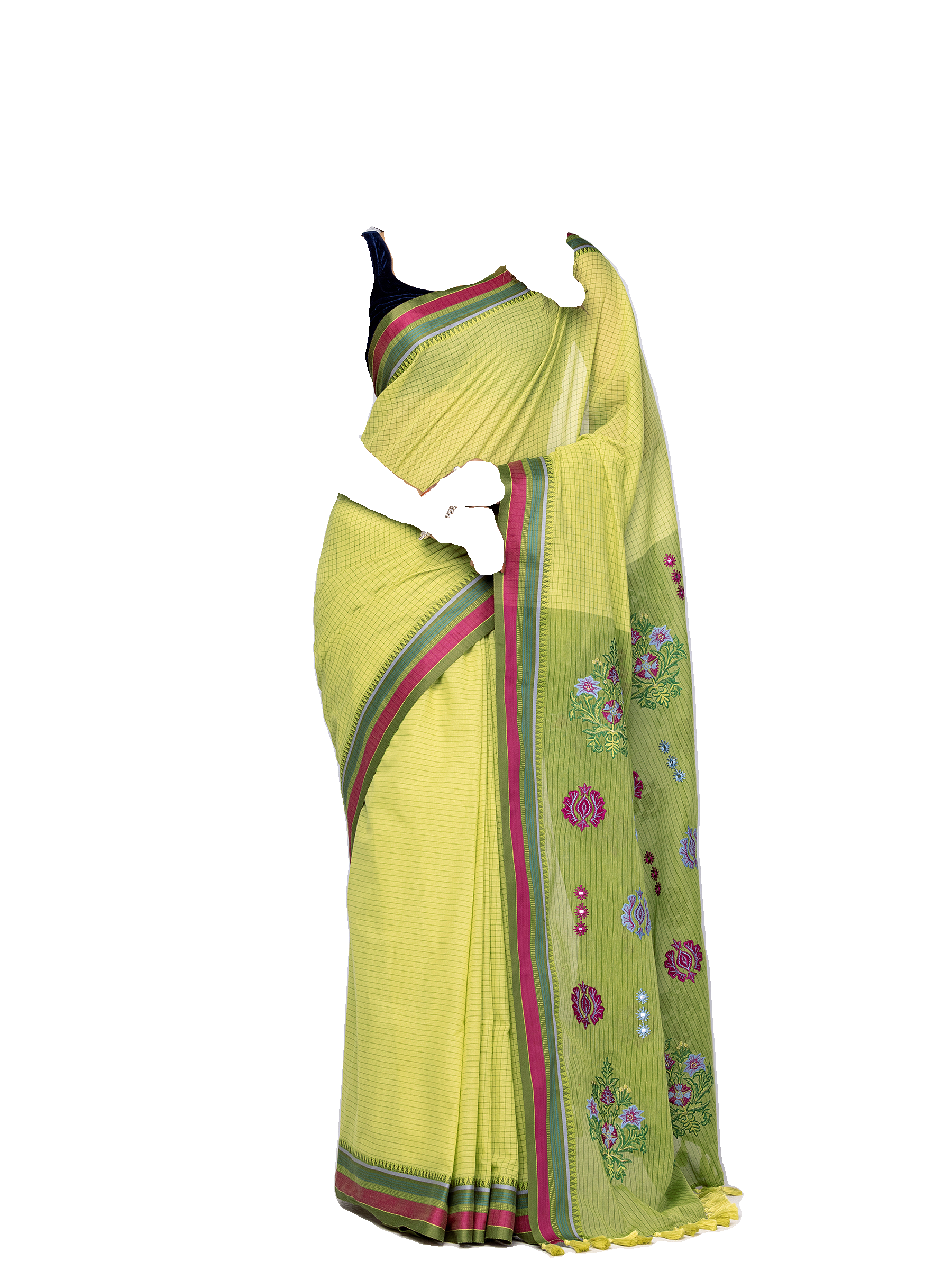}
    \end{subfigure}

    \vspace{4pt}
    \begin{subfigure}{0.19\textwidth}
        \includegraphics[width=\linewidth]{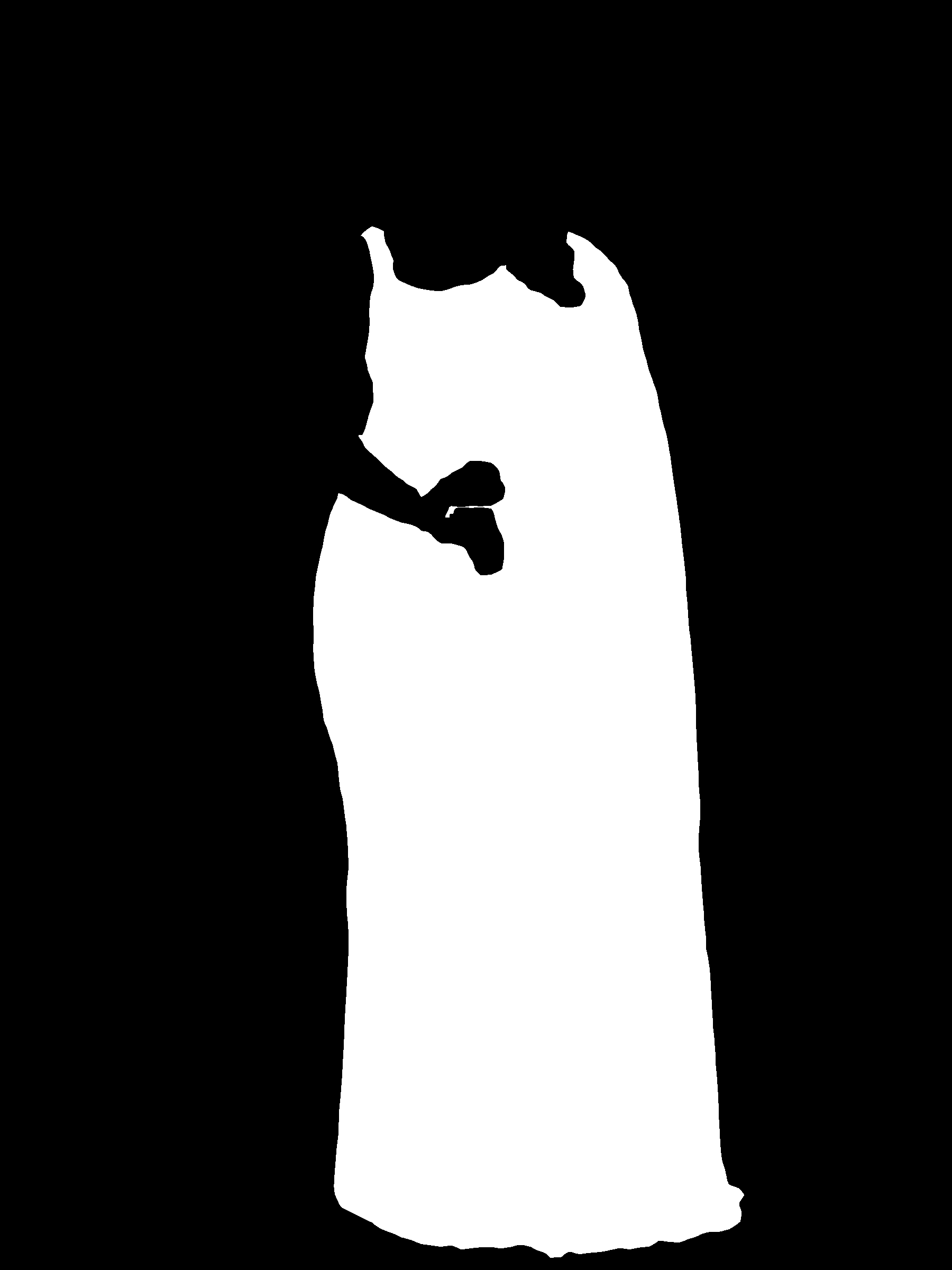}
    \end{subfigure}
    \hfill
    \begin{subfigure}{0.19\textwidth}
        \includegraphics[width=\linewidth]{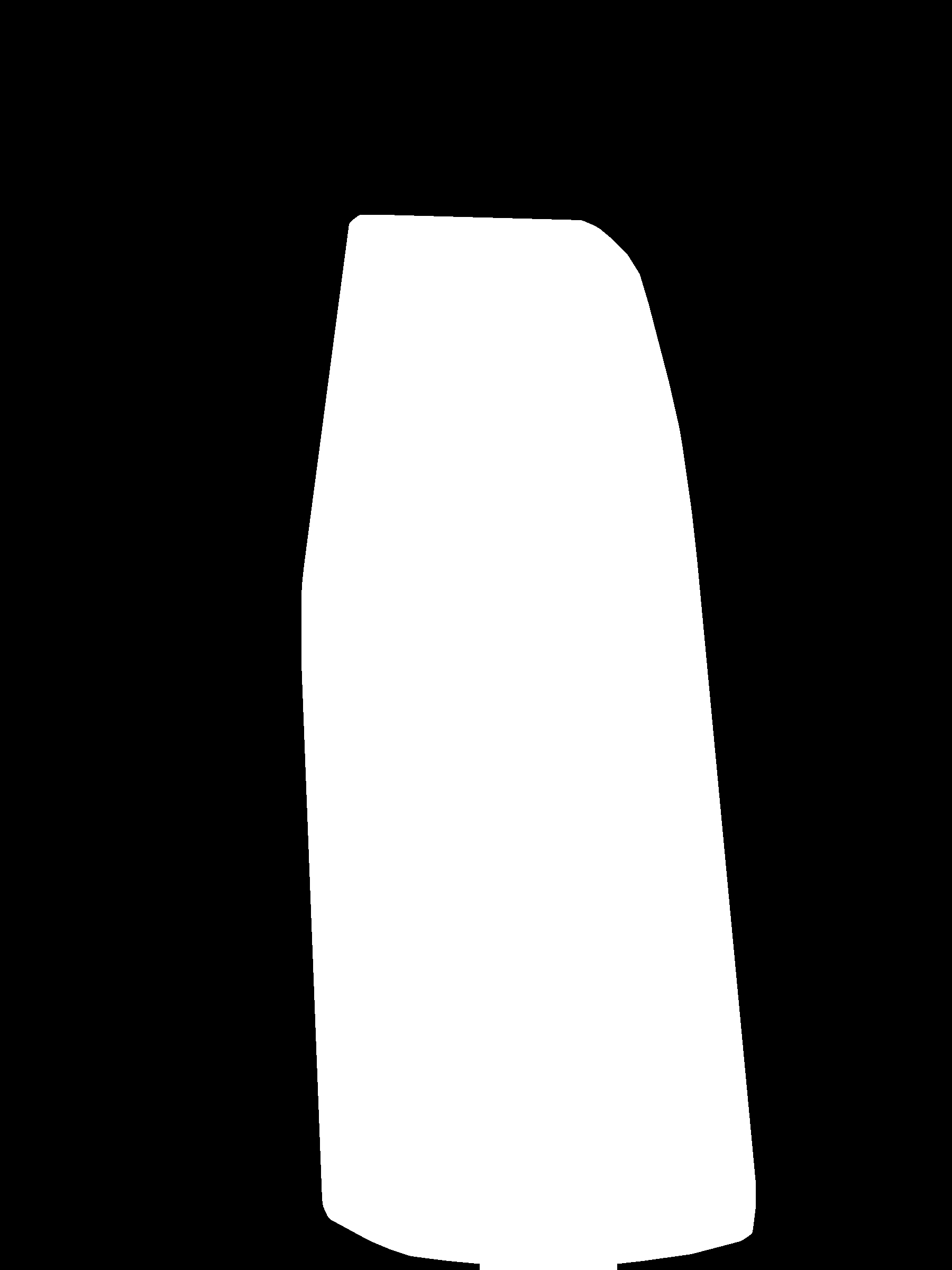}
    \end{subfigure}
    \hfill
    \begin{subfigure}{0.19\textwidth}
        \includegraphics[width=\linewidth]{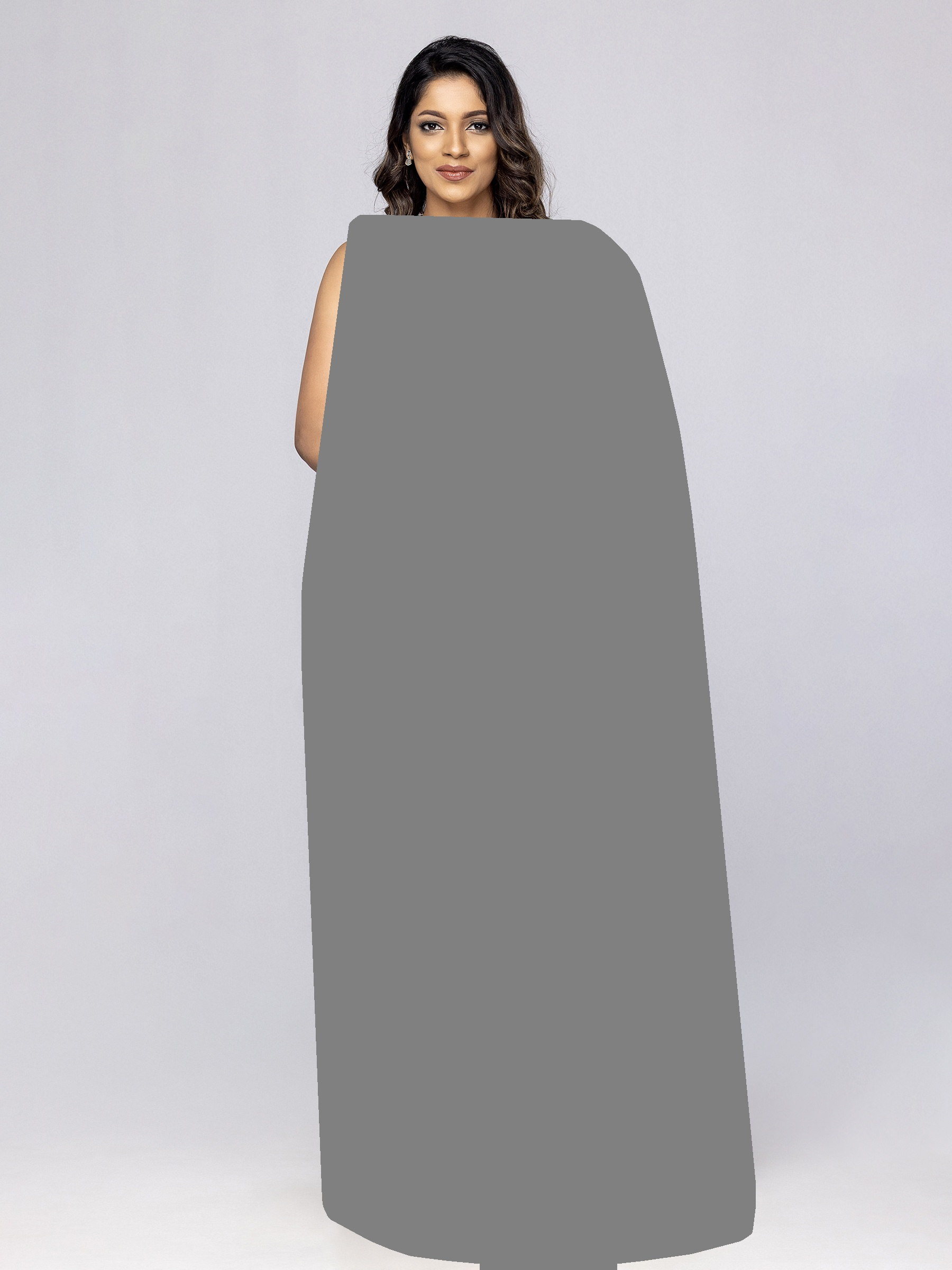}
    \end{subfigure}
    \hfill
    \begin{subfigure}{0.19\textwidth}
        \includegraphics[width=\linewidth]{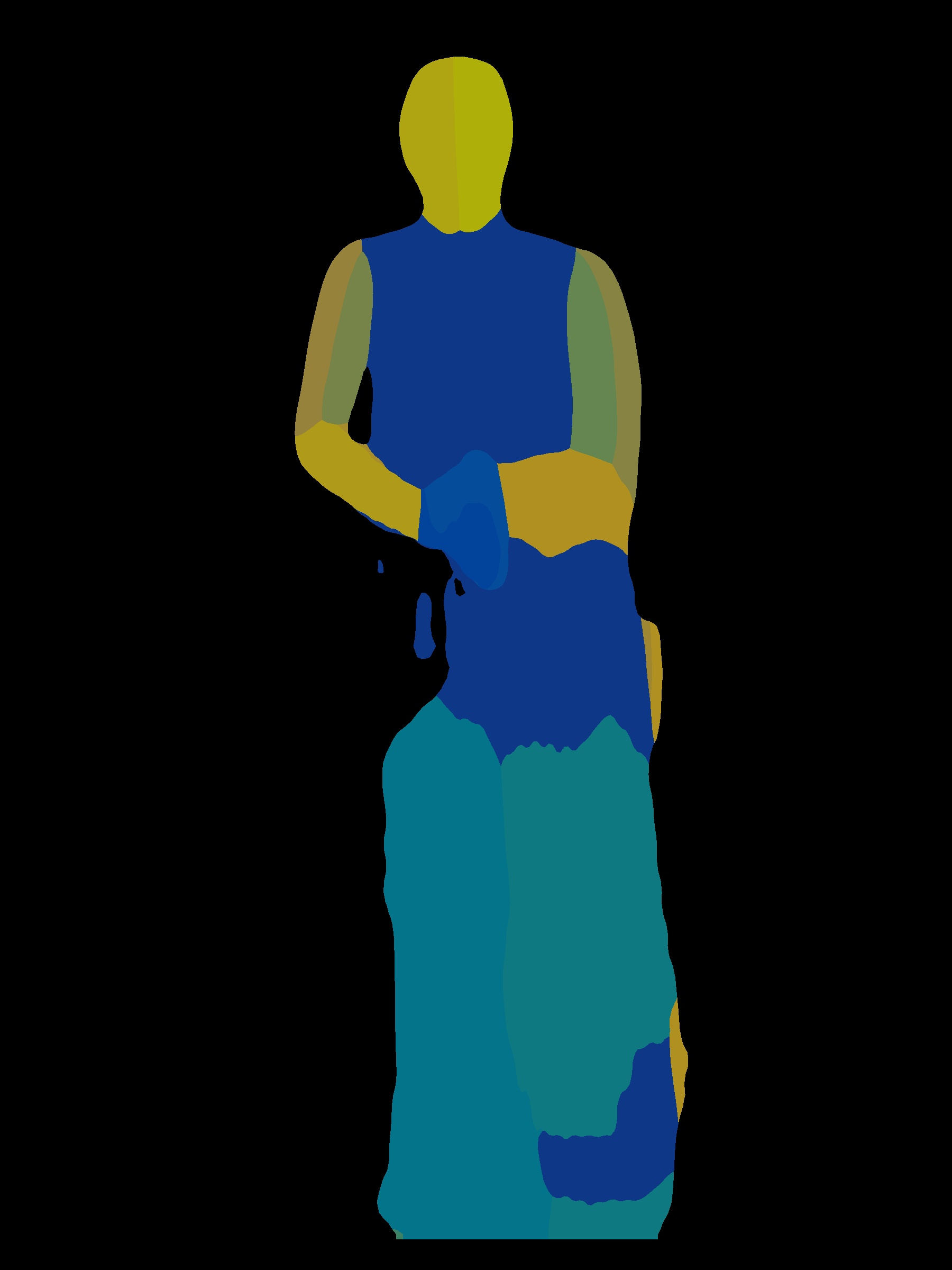}
    \end{subfigure}
    \hfill
    \begin{subfigure}{0.19\textwidth}
        \includegraphics[width=\linewidth]{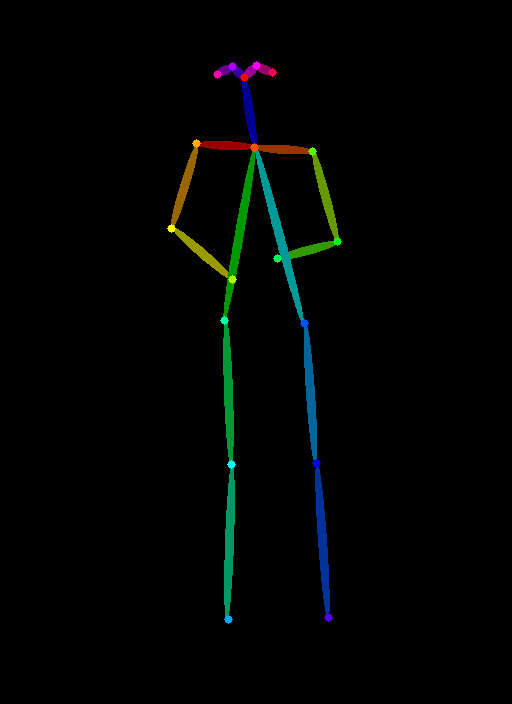}
    \end{subfigure}

    \caption{BD-VITON composition example. 
    Top row: Original Image $I_{org}$, Parse $I_{parse}$, Parse-Agnostic $I_{parseAgn}$, Warped Cloth Mask $I_{wClothM}$, Cloth $I_{cloth}$. 
    Bottom row: Cloth Mask $I_{ClothM}$, Agnostic Mask $I_{AgnM}$, Agnostic $I_{Agn}$, Densepose $I_{dp}$, OpenPose Rendering $I_{op}$.}
    \label{fig:dataset_composition}
\end{figure*}

A pretrained human parsing model (SCHP~\cite{schp} along with FASHN Human Parser~\cite{fashn}) was employed to obtain pixel-level semantic structure for different body parts. Each individual image is converted to a dense segmentation map where every pixel is assigned a semantic label depending on the body part it corresponds to. The segmented maps are then remapped to LIP-style label space, generating the parse image $I_{parse}$ as shown in Figure~\ref{fig:dataset_composition}.

These segmented maps provide region-level supervision and explicitly separate body parts from garments. The labels are also generated by the pretrained networks we have used.

\subsubsection{Parse-Agnostic Representation}
A parse-agnostic representation $I_{parseAgn}$, is constructed to prevent garment leakage during synthesis. From the semantic mask, the clothing related and arm regions are removed including Upper Garments, Dresses, Coats, Pants, etc. Arm regions are removed using pose guided masking using thick skeleton connection between keypoint which ensures complete coverage. The resulting segmentation map eliminates garment-specific pixels while preserving body structure.

The parse-agnostic maps are used as structural inputs that guide garment replacement while preventing identity copying.

\subsubsection{Cloth Mask Extraction}
We define a garment category set $\mathcal{C}_{cloth}$ consisting of upper-body, lower-body, and full-body garment labels. 
A binary cloth mask is constructed as
\begin{equation}
I_{wClothM}(x,y) =
\begin{cases}
1 & \text{if } S(x,y) \in \mathcal{C}_{cloth} \\
0 & \text{otherwise.}
\end{cases}
\end{equation}
Here $S(x,y)$ refers to the semantic label of the pixel as determined by FASHN Human Parser~\cite{fashn}.

\subsubsection{Cloth Extraction}
In addition to full-body parsing, we extract garment-only images to serve as explicit clothing inputs for the virtual try-on models. Given a semantic parsing map, we isolate clothing-related regions by selecting a set of garment class labels and constructing a binary garment mask:
\begin{equation}
I_{wClothM}(x,y) = \mathbb{1}\{(x,y)\in \mathcal{C}_{\text{cloth}}\},
\end{equation}
where $\mathbb{1}\{\cdot\}$ denotes the indicator function and $\mathcal{C}_{\text{cloth}}$ is the set of clothing-related semantic classes (e.g., upper garments, dresses, pants, scarves/dupatta).

We obtain the garment-only image by masking the original image with $I_{cloth}$:
\begin{equation}
I_{cloth} = I_{org} \odot I_{wClothM},
\end{equation}
where $\odot$ denotes element-wise multiplication (broadcast across color channels). Pixels outside the garment region are set to a constant background value (or transparency), yielding a cloth-only representation that preserves garment texture and structure while removing background and body regions. This procedure enables consistent garment conditioning for downstream synthesis.

\subsubsection{Agnostic Mask Generation}
To remove garment appearance while preserving human pose and identity cues, we construct a clothing-agnostic representation using semantic human parsing followed by mask refinement.

Given the input image $I_{org} \in \mathbb{R}^{H \times W \times 3}$, we obtain a pixel-wise semantic segmentation using a transformer-based human parsing model built upon SegFormer~\cite{xie2021segformer}. 
The network predicts a dense label map
\begin{equation}
S = \mathcal{F}(I), \quad S \in \{0,\dots,C-1\}^{H \times W},
\end{equation}
where $S(x,y)$ denotes the predicted class label at pixel $(x,y)$ and C represents the total number of semantic classes predicted by the human parsing model.

Since this model is also trained to perform optimally for western apparel, we adhered to structural abstraction that enables generalization to traditional Bangladeshi attire such as \emph{saree}, \emph{panjabi}, and \emph{salwar kameez}, which are mapped to semantically closest garment classes based on spatial coverage.

Since draped structures and layered fabric in more complex garments like saree exist, raw segmentation masks may contain fragmented components. 
We therefore apply morphological closing, connected-component filtering, and scale-adaptive dilation to ensure complete garment coverage. 
To suppress garment-specific structural details such as folds and concave boundaries, we compute the convex hull over the garment region, producing a smooth unified silhouette denoted as $I_{AgnM}$.

The final agnostic image $I_{Agn}$ is obtained by replacing garment pixels with a constant neutral value $c$:
\begin{equation}
I_{Agn}(x,y) =
\begin{cases}
c & \text{if } I_{AgnM}(x,y) = 1 \\
I_{org}(x,y) & \text{otherwise.}
\end{cases}
\end{equation}
This representation removes fabric texture, color, and stylistic identity while preserving pose, facial attributes, and background context. 
The abstraction operates at a structural garment level, enabling robust application to culturally specific attire without requiring garment-specific retraining.

\subsubsection{Human Pose Estimation}
A pretrained DWPose model~\cite{dwpose} was used to extract 2D body keypoints in the OpenPose COCO-18 format~\cite{openpose}. For every image, the DWPose model predicts 18 anatomical points corresponding to the major joints of the body (e.g, shoulders, knee, elbow etc) and is represented by ($x_i$, $y_i$, $c_i$) where $c_i$ represents confidence.

These pose annotation are then stored and used as structural conditioning inputs for garment wrapping. Rendered pose maps were also generated for visualization and debugging.

The pose map is used to encode the 2D joint locations of the person and provides explicit structural information to the model. The model then uses this information to interpret the target pose when forming the person-agnostic representation, or more specifically it identifies where the garment region should be synthesized. Pose map conditioning also helps the warping stage align the input cloth with the body layout.

\begin{table}[t]
\centering
\caption{Pose Diversity of BD-VITON}
\begin{tabular}{lcc}
\toprule
\textbf{Pose Attribute} & \textbf{Value} & \textbf{Remarks} \\
\midrule
Frontal Pose & 87.16\% & Subject facing camera \\
Slightly Turned Pose & 12.84\% & Minor body rotation \\
Avg. Visible Keypoints & 17.62/18 & Confidence $\geq$ 0.2 \\
\bottomrule
\end{tabular}
\end{table}

\subsubsection{DensePose Map Generation}
Dense pose estimation provides pixel-to-surface correspondence between 2D images and 3D body models. We employed DensePose-RCNN\cite{guler2018densepose} with ResNet-50-FPN backbone which combines Faster R-CNN\cite{ren2015faster} for person detection with a fully convolutional head for dense correspondence prediction. The model outputs 24-part body segmentation $P \in \{0,...,24\} ^ {H\times W}$ following the SMPL model\cite{loper2015smpl} and UV coordinates $(U,V) \in [0,1]^{H\times W\times 3}$ representing surface parameterization within each part. For multiple detected persons, we select the largest bounding box as the primary subject. The final DensePose map represented as $I_{dp} \in R^{H\times W\times 3}$ is rendered as color-coded 24-part segmentation where
\begin{equation}
I_{dp}(x,y)=
\begin{cases}
\operatorname{ColorMap}\!\big(P(x,y)\big), & \text{if } (x,y)\in \Omega_{\mathrm{det}}\\
(0,0,0)^{\top} & \text{otherwise}
\end{cases}
\end{equation}
which in turn provides complete surface topology to guide garment synthesis beyond skeletal pose representations.

DensePose images are necessary because they provide dense pixel-level body surface correspondence that guides accurate garment warping, pose-aware alignment and high-resolution realistic synthesis.

\section{Methods}
\label{sec:blind}
Given a person image $I$ and a clothing image $C$, a framework designed to solve Virtual Try-On task should generate an image $\hat{I}$ that depicts the same person wearing the garment $C$. Although this formulation appears intuitive, collecting such triplets $\{I,C,\hat{I}\}$ is non-trivial. Therefore, a common strategy is to first construct a clothing-agnostic representation of the person image, denoted as $I_a$, which removes clothing information from $I$ to prevent it from influencing the model. After obtaining $I_a$, the framework leverages $I_a$ together with the clothing image $C$ to generate the target image. Since the original image $I$ is already available, the generated output $I_g$ can be compared with $I$, allowing the framework to be trained in a paired manner. All three methods used in this study, VITON-HD~\cite{choi2021vitonhd}, HR-VITON~\cite{lee2022hrviton}, and StableVITON~\cite{kim2023stableviton}, follow this general paradigm. However, their architectures have progressively become more sophisticated and require additional inputs, such as dense pose maps, to address the complex challenges inherent in virtual try-on tasks. To satisfy these input requirements, we preprocessed the samples and generated all necessary inputs for applying these methods to the BD-VITON dataset. In the following subsections, we provide a brief overview of these methods, along with the approaches we employed to adapt them to BD-VITON.

\subsection{VITON-HD}

\subsubsection{Overview}
VITON-HD is a high-resolution virtual try-on framework\cite{choi2021vitonhd} that decomposes garment transfer into three stages: semantic layout prediction, geometric alignment and photorealistic synthesis. The segmentation generator uses a U-Net architecture\cite{ronneberger2015unet} with clothing-agnostic person representation, pose keypoints and target garment to predict human parsing maps via cross-entropy loss. The geometric matching module warps the garment using Thin-Plate Spline transformations\cite{bookstein1989tps} to match the target pose and body shape. The ALIAS (ALIgnment-Aware Segment) generator synthesizes the final output through alignment-aware normalization that independently normalizes aligned and misaligned feature regions combined with perceptual\cite{johnson2016perceptual} and adversarial losses\cite{goodfellow2014gan,isola2017pix2pix}.

\subsubsection{Adaptation}
Although the inference script and pretrained checkpoints were available, the training script for VITON-HD was not provided. Therefore, we implemented our own training script, adopting a transfer learning approach by initializing all three stages with the official pretrained checkpoints. Moreover, we strictly follow the loss functions provided in VITON-HD in training. We have used Eq.~\ref{eq:vitonhd_seg_loss} to compute the loss for the segmentation generator.
\begin{equation}
\mathcal{L}_{S}=\mathcal{L}_{\mathrm{cGAN}}+\lambda_{\mathrm{CE}}\mathcal{L}_{\mathrm{CE}} .
\label{eq:vitonhd_seg_loss}
\end{equation}
Here $L_S$ denotes the total loss of the segmentation generator whereas $L_{CE}$ and $L_{cGAN}$ denotes the pixel-wise cross-entropy loss and conditional adversarial loss between the segmentation map and the synthetic segmentation generated by the generator. Subsequently, for the GMM module, the loss is computed as the L1 reconstruction error between the predicted warped garment and the ground-truth aligned reference. For the ALIAS generator, a multi-component loss function is utilized, where we have L1 reconstruction loss for pixel-level accuracy, VGG perceptual loss\cite{johnson2016perceptual} to capture semantic similarity and adversarial GAN loss\cite{goodfellow2014gan} for generating realistic images.

\subsection{HR-VITON}
\subsubsection{Overview}
HR-VITON~\cite{lee2022hrviton} adopts a comparatively simpler architectural design consisting of two stages: the Condition Generator and the Image Generator. In the first stage, the condition generator (TOCG) predicts both the synthetic segmentation map and the representation of the warped cloth, effectively combining the first two stages of VITON-HD into a single module. In the second stage, the image generator takes the warped cloth, pose map, and clothing-agnostic person representation as inputs and synthesizes the final try-on image by blending the garment onto the target person.

\subsubsection{Adaptation}
For BD-VITON, cloth images are obtained by performing simple garment extraction from the person images, thereby creating pseudo-cloth images. As a result, during the condition generator stage, HR-VITON tends to learn that directly copying and pasting the cloth image onto the person image is the easiest way to minimize the loss. This behavior leads to undesirable outcomes, particularly during unpaired testing, and produces nearly identical results in paired testing.

To mitigate this issue, we first introduced Gaussian noise along the edges of the extracted garments to make them visually distinct from the original clothing regions in the person image. In addition, we generated multiple augmented versions of each cloth image by applying rotations. A new training set was then created using these augmented samples, where a single person image is paired with multiple augmented cloth images, better approximating the characteristics of real standalone garment images. With the help of these two strategies, the condition generator stage was able to produce meaningful warped garment outputs, without simply copying and pasting the cloth image onto the person. 

\subsection{StableVITON}
\subsubsection{Overview} StableVITON is a diffusion-based virtual try-on framework\cite{kim2023stableviton} built on a pre-trained latent diffusion backbone\cite{rombach2022highresolution} that formulates garment transfer as conditional inpainting in latent space. Given a clothing-agnostic person representation, an agnostic clothing mask, and DensePose guidance\cite{guler2018densepose}, the denoising U-Net is modified to accept these conditions alongside the noisy latent and iteratively synthesizes the try-on result. To inject garment appearance, StableVITON uses a CLIP image encoder\cite{radford2021clip} for global exemplar conditioning and a dedicated spatial garment encoder whose features interact with U-Net features via zero cross-attention blocks, allowing the model to learn implicit alignment between the target garment and the body directly during diffusion-based generation~\cite{kim2023stableviton}.

Unlike the former two models, StableVITON did not require any fundamental adaptation for BD-VITON.

\subsection{Experimental Setup}
The trainings were performed using an RTX 5060 Ti GPU. Due to its Blackwell architecture, the recommended CUDA version in the code base was incompatible with our setup, so we switched to CUDA 13.0 to train the model. Consequently, the xFormers~\cite{xFormers2022} module was not compatible with this CUDA version, for which we had to use Scaled Dot Product Attention transformer module (SDPA) which is built into this GPU. All models were trained at a resolution of $512\times384$.

For VITON-HD~\cite{choi2021vitonhd}, the Segmentation Generator and the Geometric Matching Module has utilized a batch size of 2 and the ALIAS Generator used a batch size of 4.
 first stage of HR-VITON~\cite{lee2022hrviton} was trained with a batch size of 4 while the second stage was trained with a batch size of 1.
For StableVITON~\cite{kim2023stableviton}, the training was performed with a batch size of 4.

\section{Results}
\subsubsection{Evaluation Metrics} We employ two complementary metrics for evaluating paired image quality and one metric for unpaired evaluation. The Structural Similarity Index (SSIM) \cite{wang2004image} measures the perceptual similarity between the generated image and the ground-truth image by comparing luminance, contrast, and structural information, where higher values indicate better structural fidelity. Learned Perceptual Image Patch Similarity (LPIPS) \cite{zhang2018unreasonable} computes perceptual distance using deep features extracted from a pre-trained VGG network, with lower values indicating better perceptual similarity and realism. For unpaired evaluation, we use the Fr\'{e}chet Inception Distance (FID) \cite{heusel2017gans}, which measures the distributional distance between generated images and real images using features extracted from a pre-trained Inception network; lower FID scores indicate that the generated images are more similar to real images in terms of visual quality and diversity.

\subsubsection{Quantitative Results} In Table \ref{tab:viton_results}, you can see with training all the methods have done better than their performance in Zero Shot. Thus, this establishes that, BD-VITON can be utilized by standard virtual try-on systems to do well for cultural specific complex structured clothing.

Although StableVITON is the most recent architecture among the evaluated methods, it achieves the lowest performance after training. This is likely due to the limited training in our experiments, as the model was fine-tuned for only five epochs because of time and resource constraints. The presence of attention layers, which typically require longer training to converge, may have further contributed to the lower performance.

Comparing HR-VITON and VITON-HD, both models achieve similar performance after training. However, their zero-shot results differ significantly, with VITON-HD performing substantially better than HR-VITON across all evaluation metrics. This discrepancy arises mainly from the behavior of the HR-VITON's condition generator, which tends to copy and paste the cloth image onto the person image when pseudo-cloth inputs are provided. Although this issue was mitigated during training through our augmentation strategies, it remains present in the zero-shot setting. As a result, HR-VITON exhibits significantly poorer zero-shot performance, particularly in terms of FID, where its score is nearly five times worse than that of StableVITON.

\begin{table}[H]
\centering
\small
\setlength{\tabcolsep}{4pt}
\begin{tabular}{l c c c c}
\hline
Model & Zero-Shot & SSIM $\uparrow$ & LPIPS $\downarrow$ & FID $\downarrow$ \\
\hline
HR-VITON & \ding{51} & 0.478 & 0.675 & 259.25 \\
HR-VITON & \ding{55} & \textbf{0.815} & \textbf{0.156} & \textbf{42.92} \\
\hline
StableVITON & \ding{51} & 0.722 & 0.238 & 69.59 \\
StableVITON & \ding{55} & \textbf{0.732} & \textbf{0.219} & \textbf{50.40} \\
\hline
VITON-HD & \ding{51} & 0.813 & 0.244 & 100.07 \\
VITON-HD & \ding{55} & \textbf{0.868} & \textbf{0.154} & \textbf{49.89} \\

\hline
\end{tabular}
\caption{Quantitative Results on BD-VITON's test set (203 images).}
\label{tab:viton_results}
\end{table}

\subsection{Qualitative Results}
\FloatBarrier
\begin{figure*}[htbp]
    \centering
    \setlength{\tabcolsep}{2pt}
    \begin{tabular}{cc|cc|cc|cc}
        \includegraphics[width=0.11\textwidth]{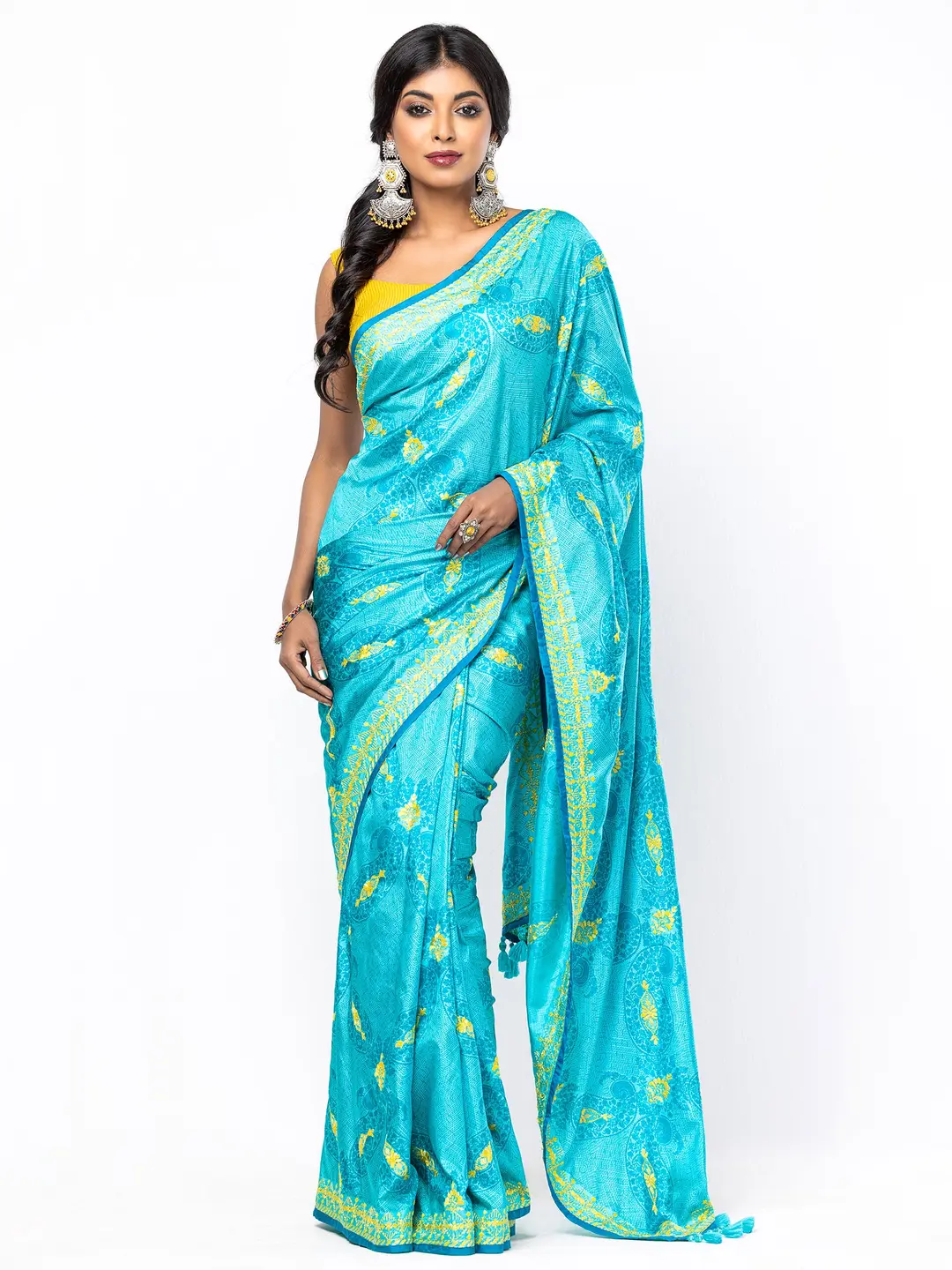} &
        \includegraphics[width=0.11\textwidth]{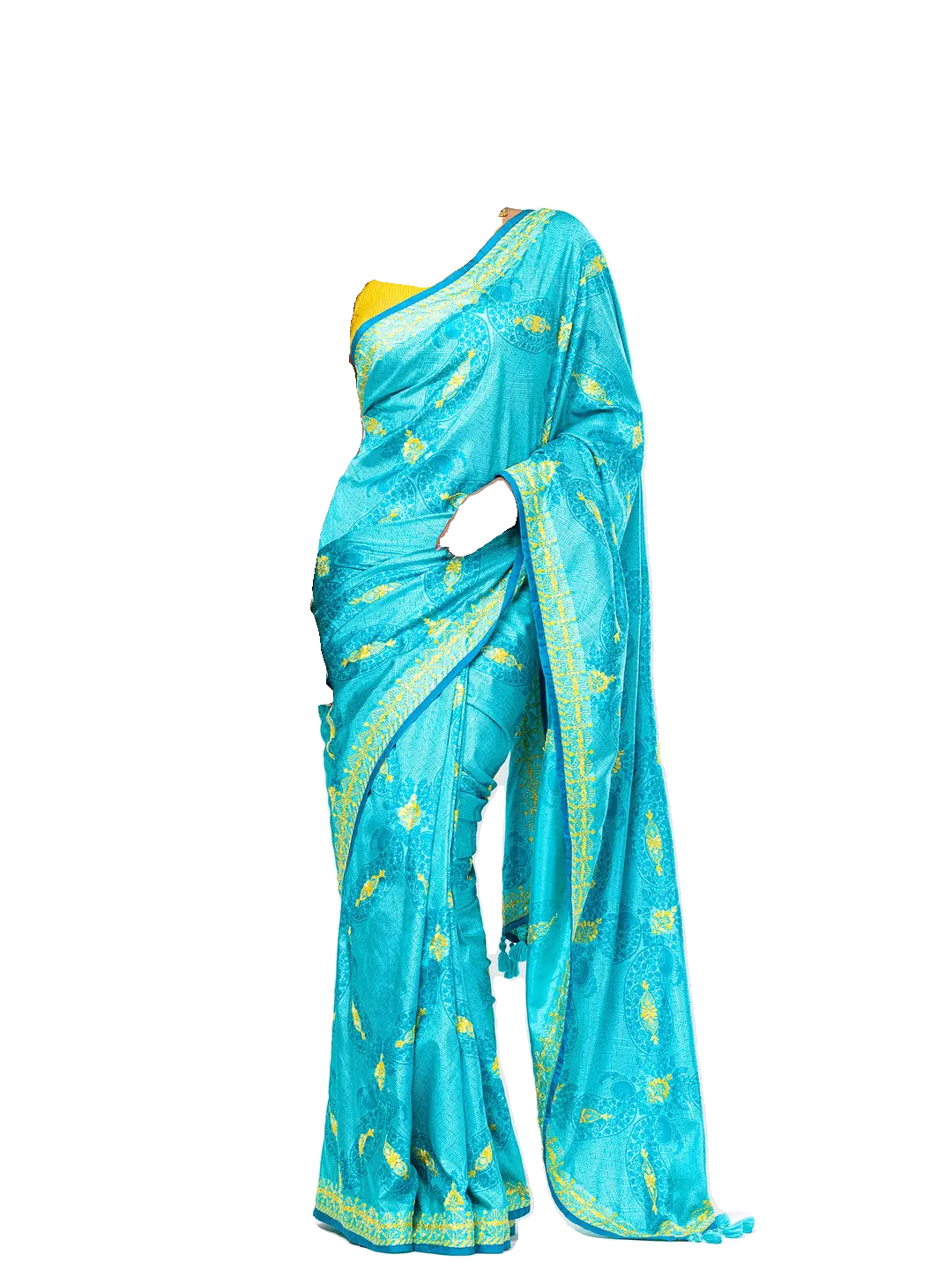} &
        \includegraphics[width=0.11\textwidth]{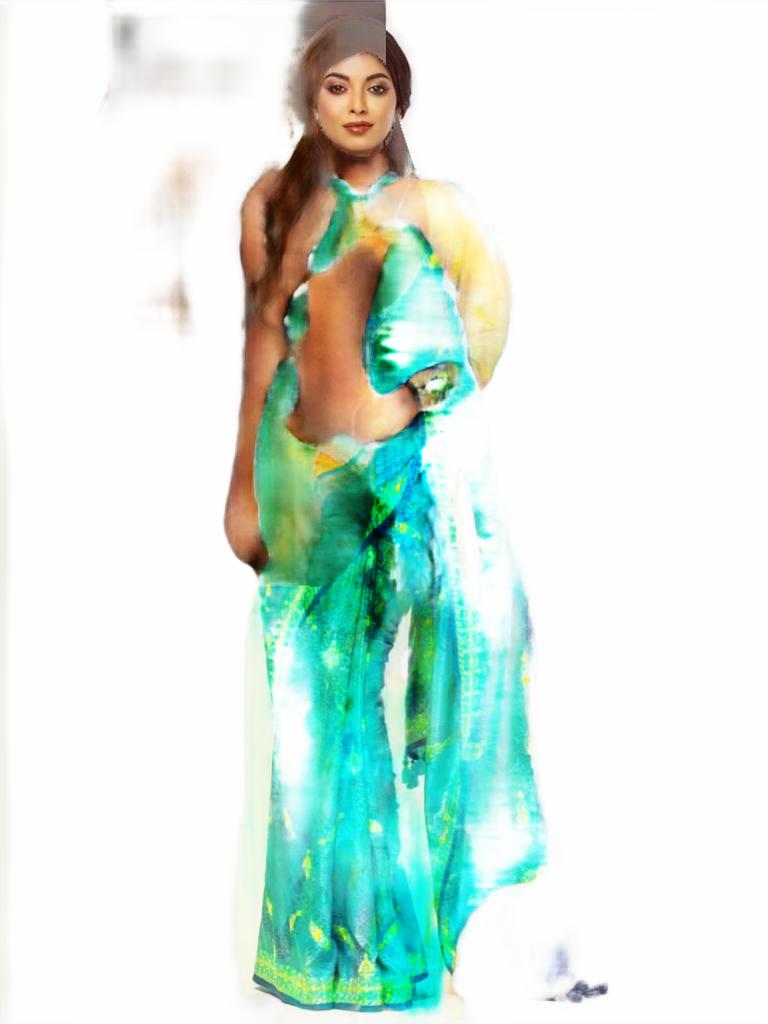} &
        \includegraphics[width=0.11\textwidth]{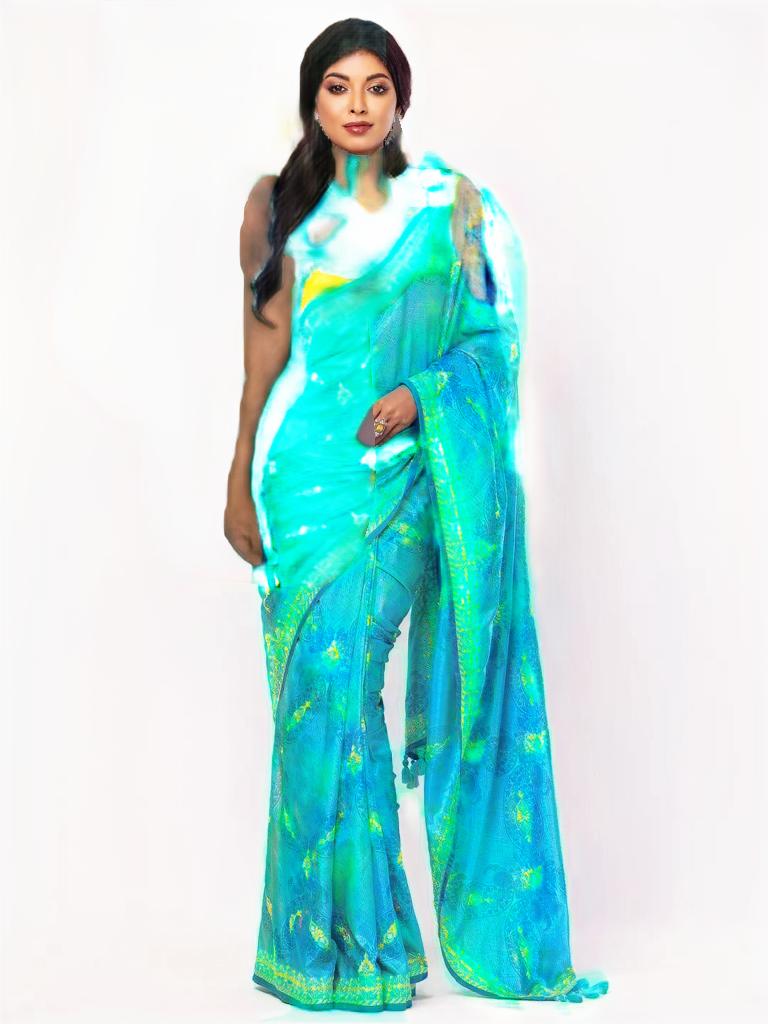} &
        \includegraphics[width=0.11\textwidth]{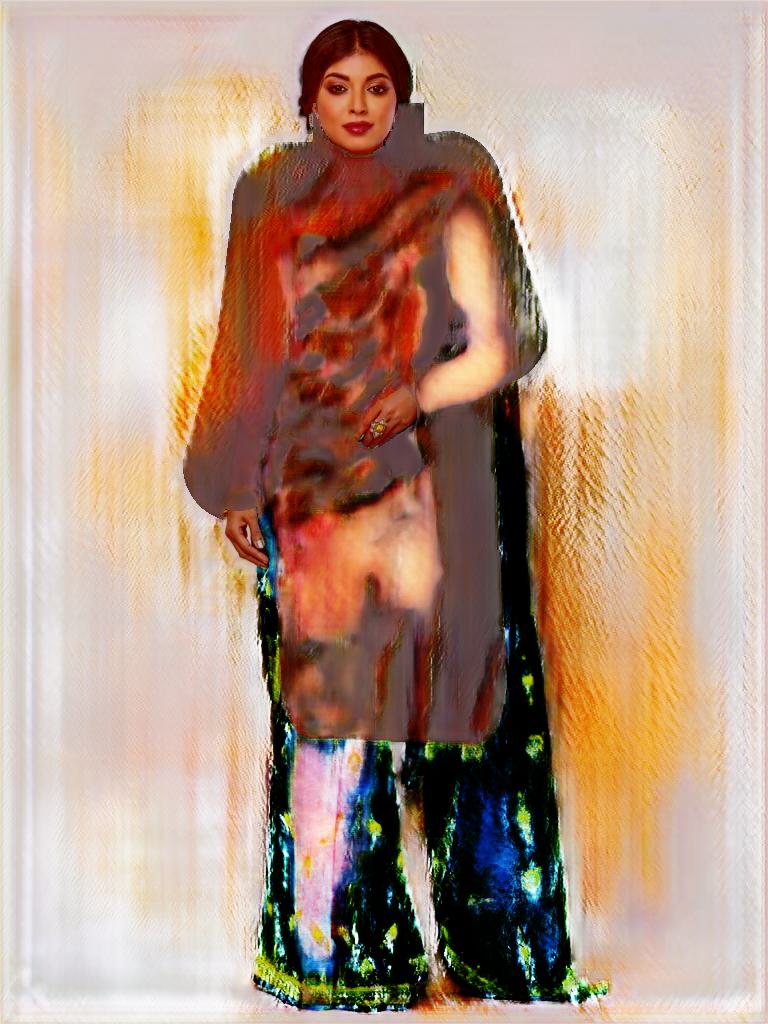} &
        \includegraphics[width=0.11\textwidth]{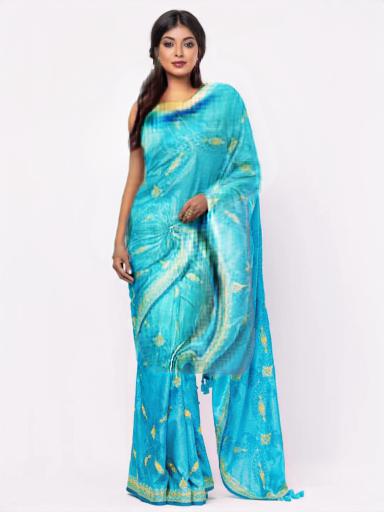} &
        \includegraphics[width=0.11\textwidth]{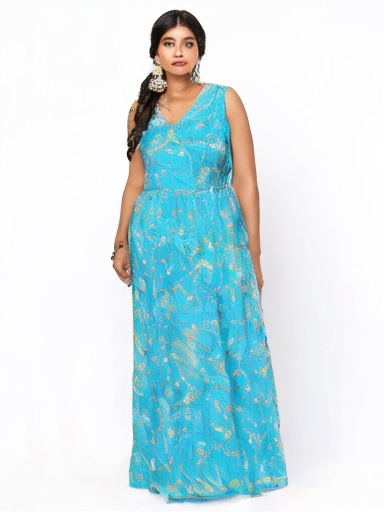} &
        \includegraphics[width=0.11\textwidth]{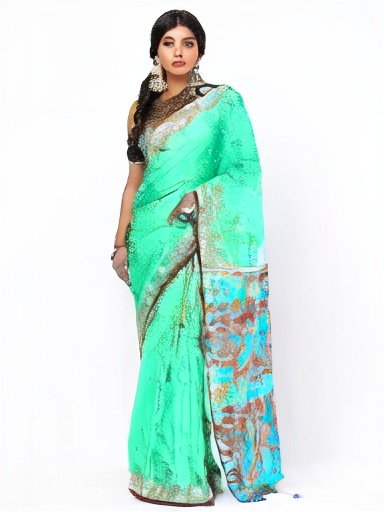} \\[4pt]

        \includegraphics[width=0.11\textwidth]{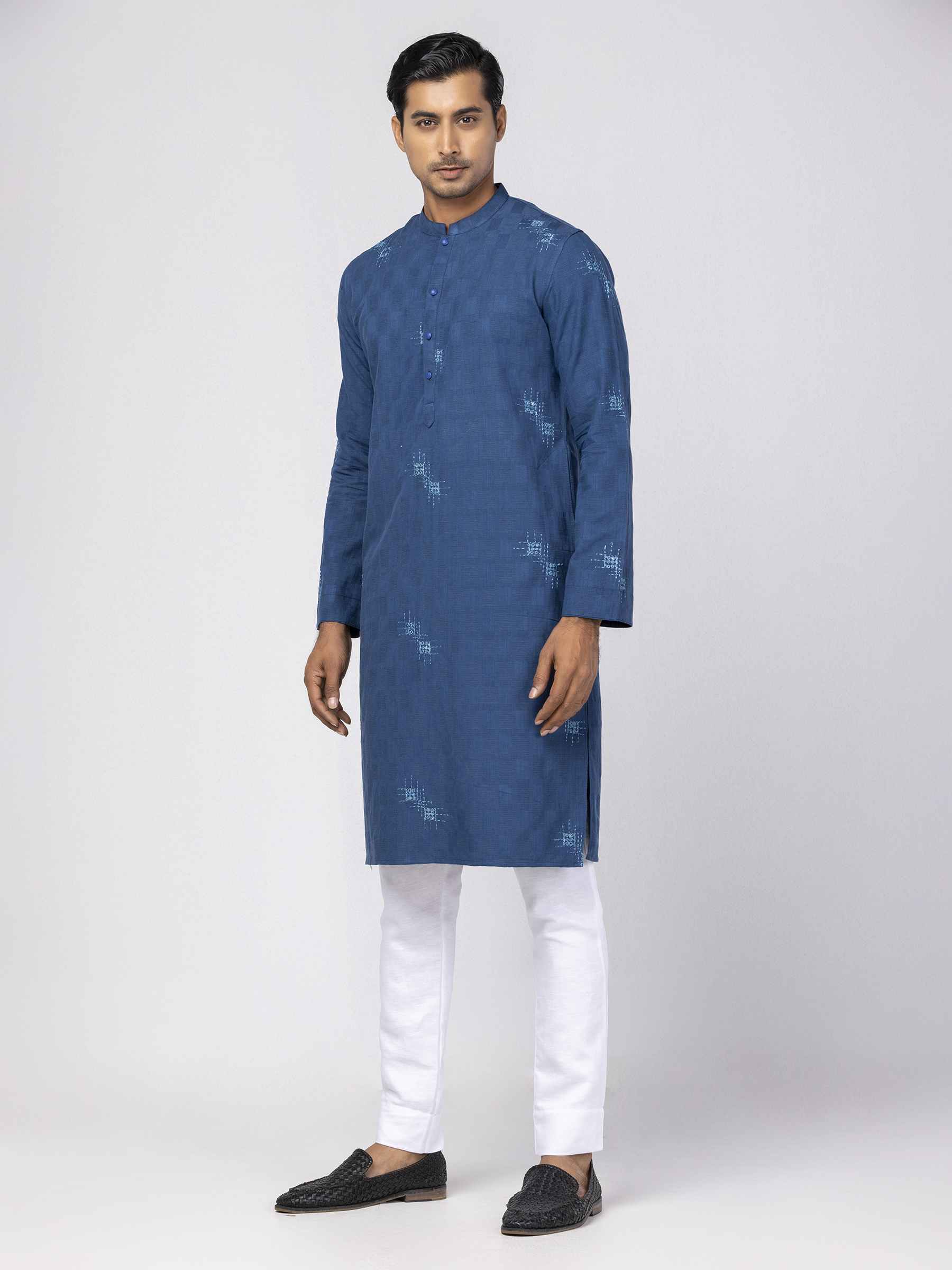} &
        \includegraphics[width=0.11\textwidth]{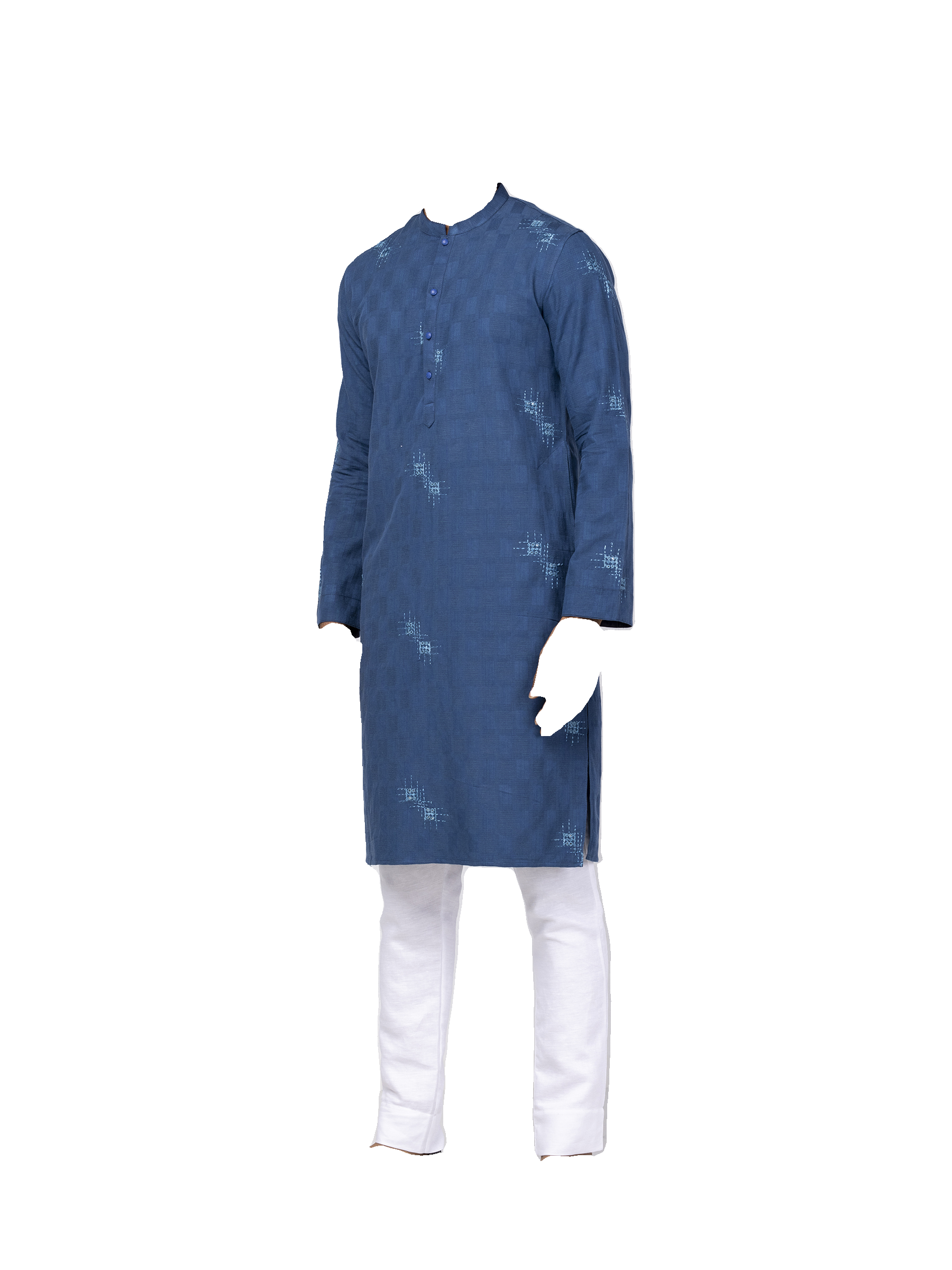} &
        \includegraphics[width=0.11\textwidth]{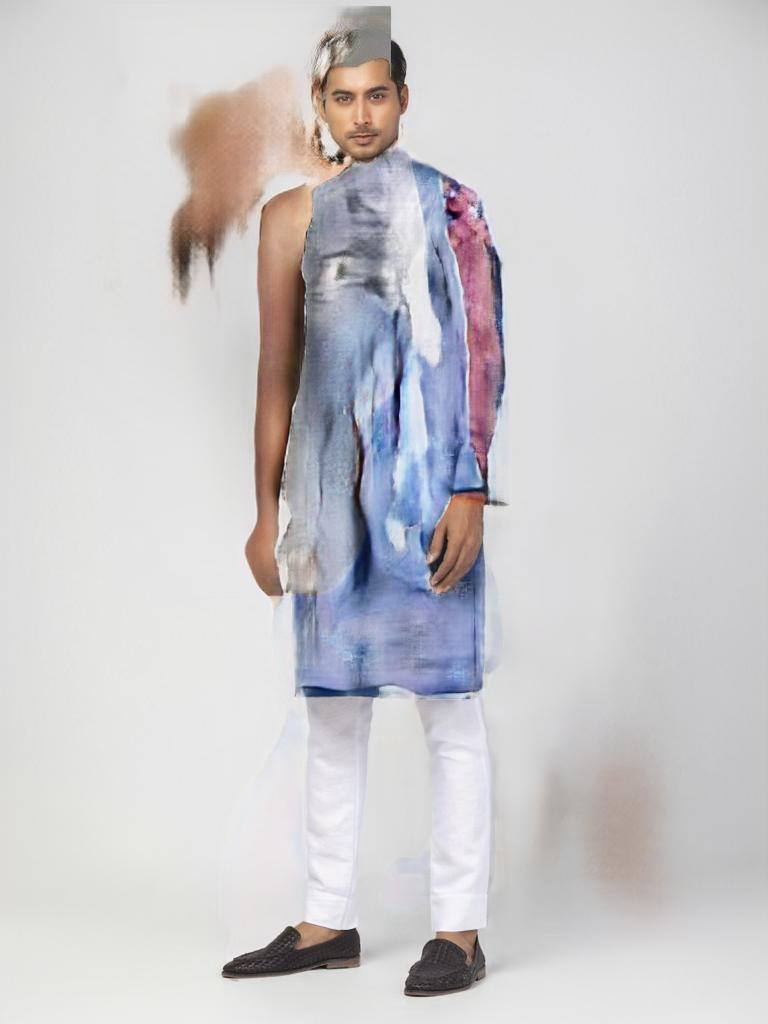} &
        \includegraphics[width=0.11\textwidth]{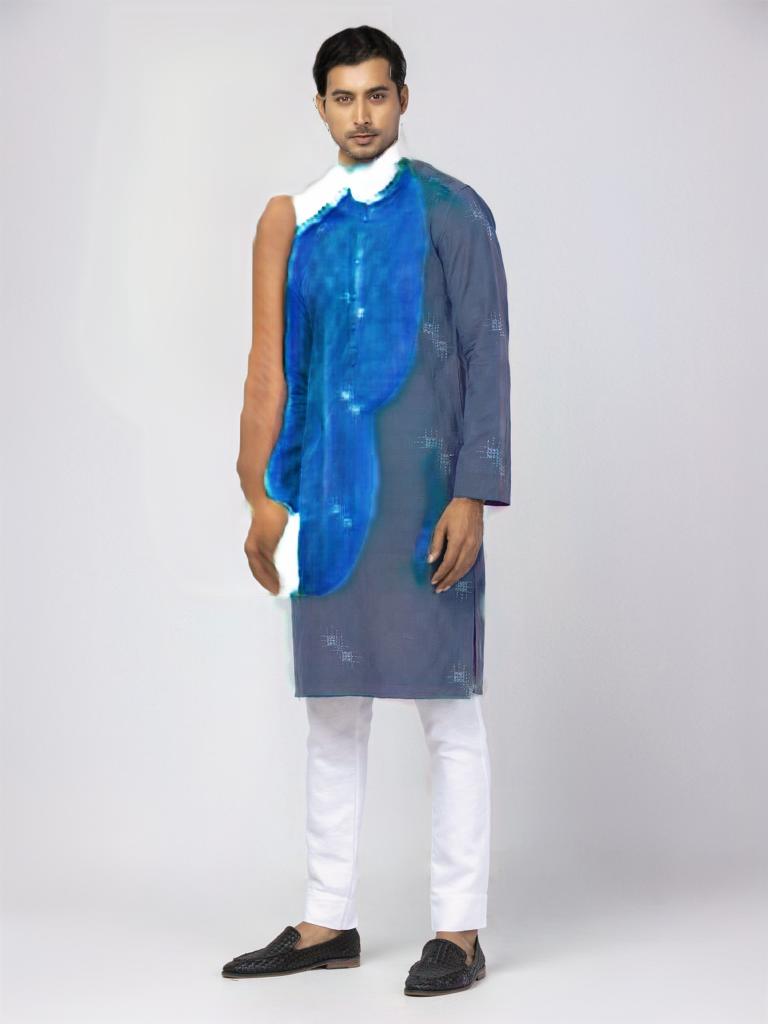} &
        \includegraphics[width=0.11\textwidth]{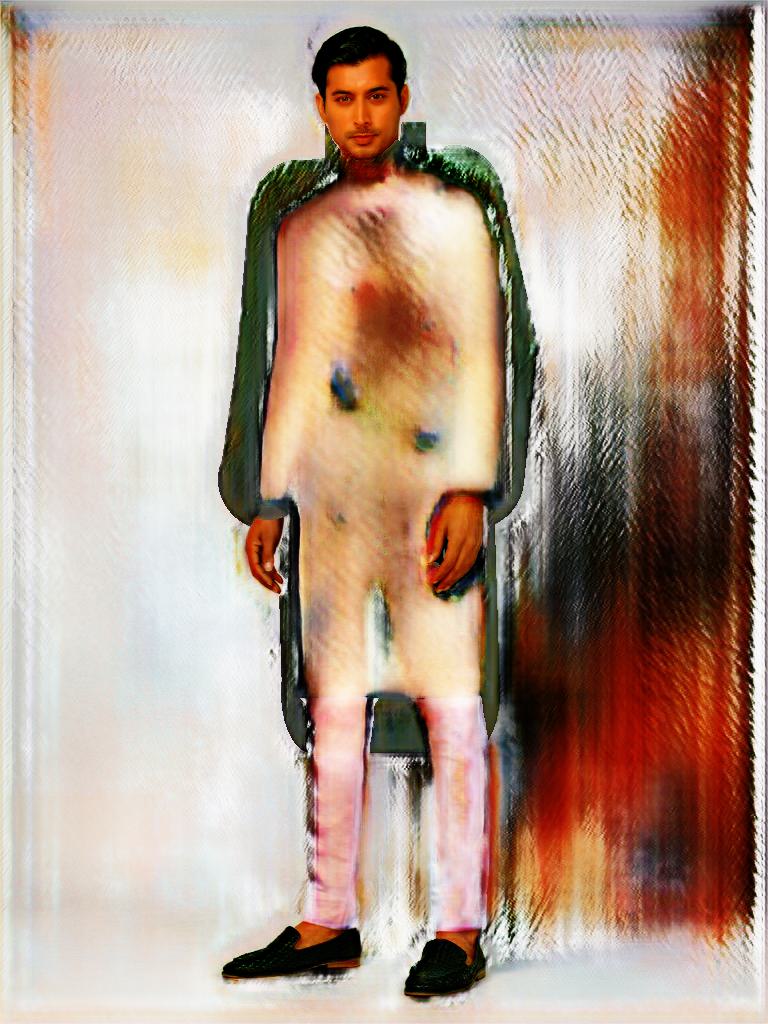} &
        \includegraphics[width=0.11\textwidth]{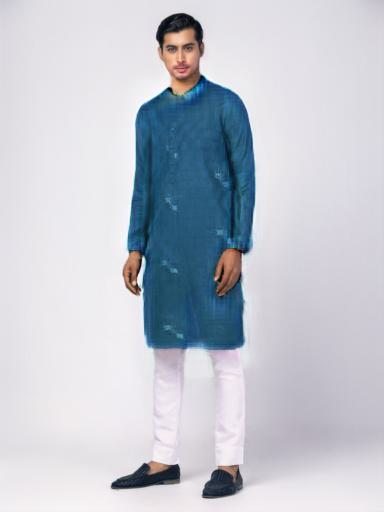} &
        \includegraphics[width=0.11\textwidth]{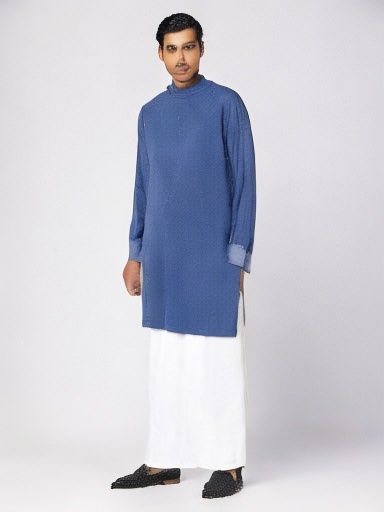} &
        \includegraphics[width=0.11\textwidth]{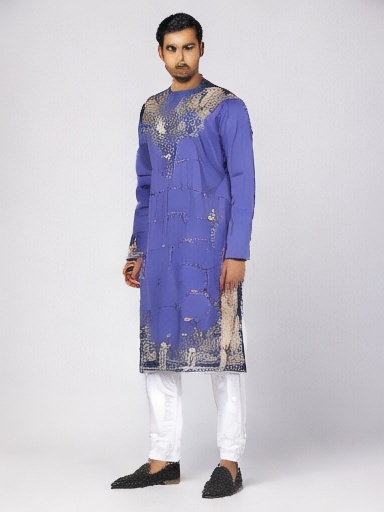} \\[4pt]

        \includegraphics[width=0.11\textwidth]{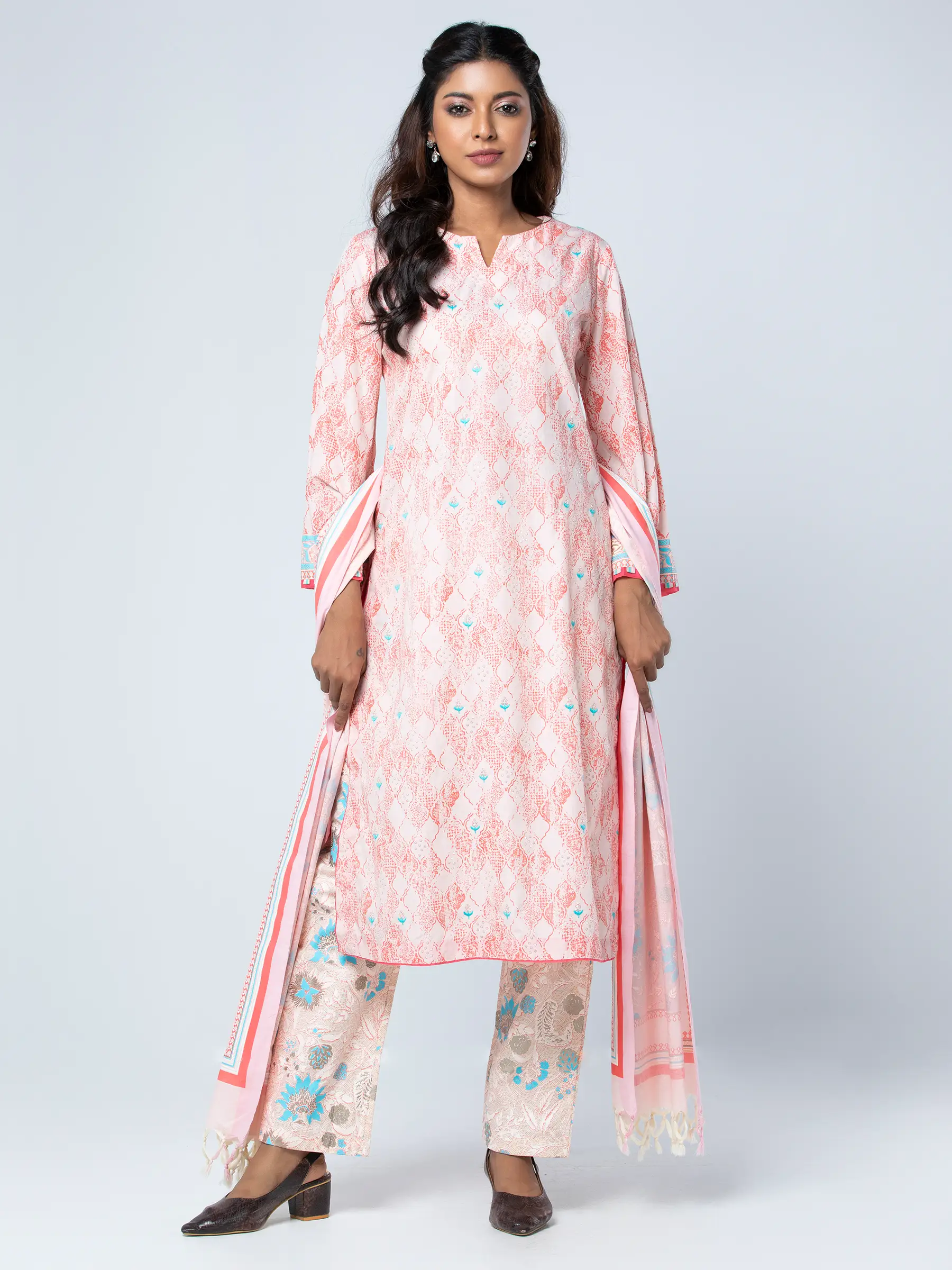} &
        \includegraphics[width=0.11\textwidth]{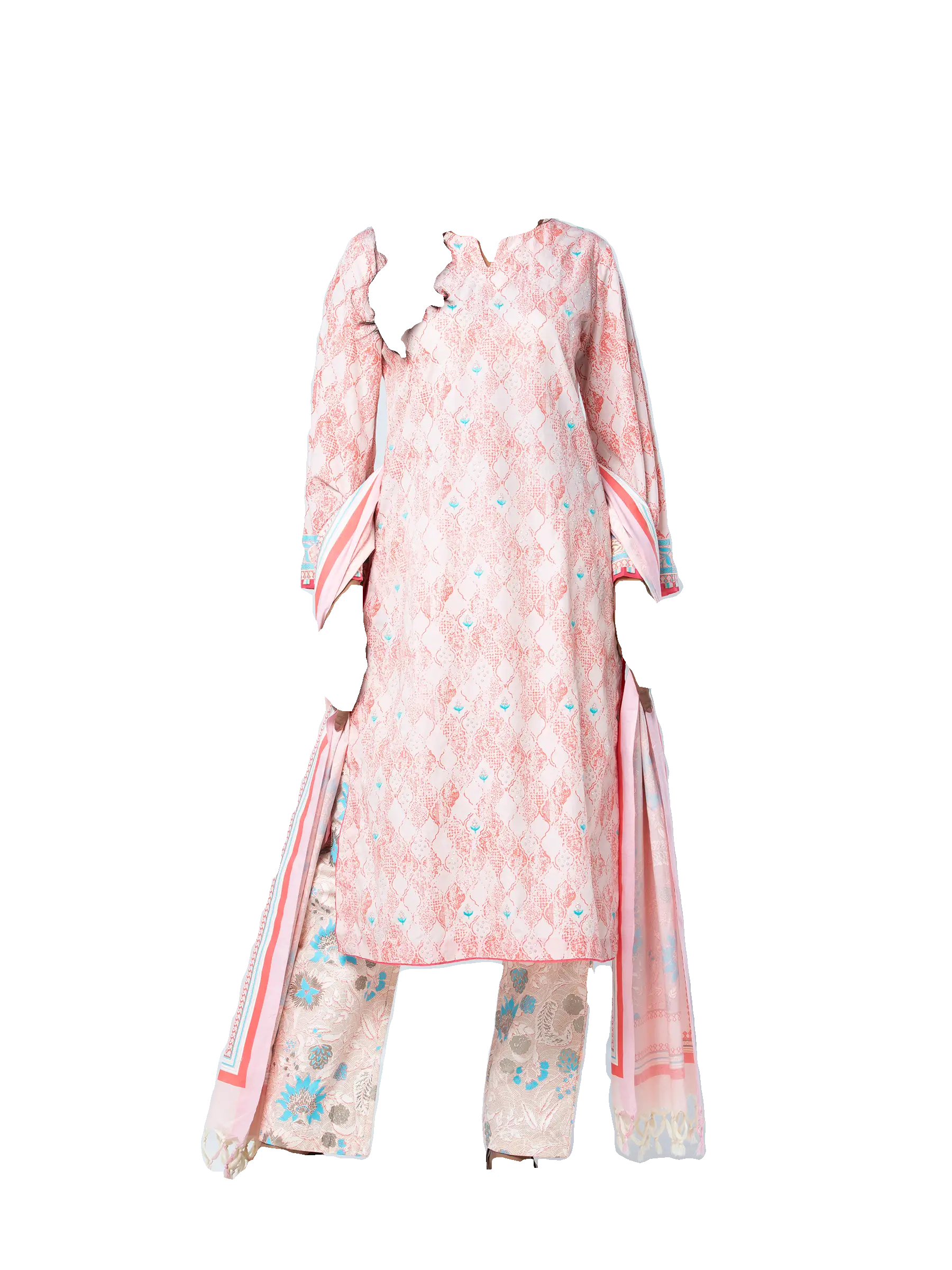} &
        \includegraphics[width=0.11\textwidth]{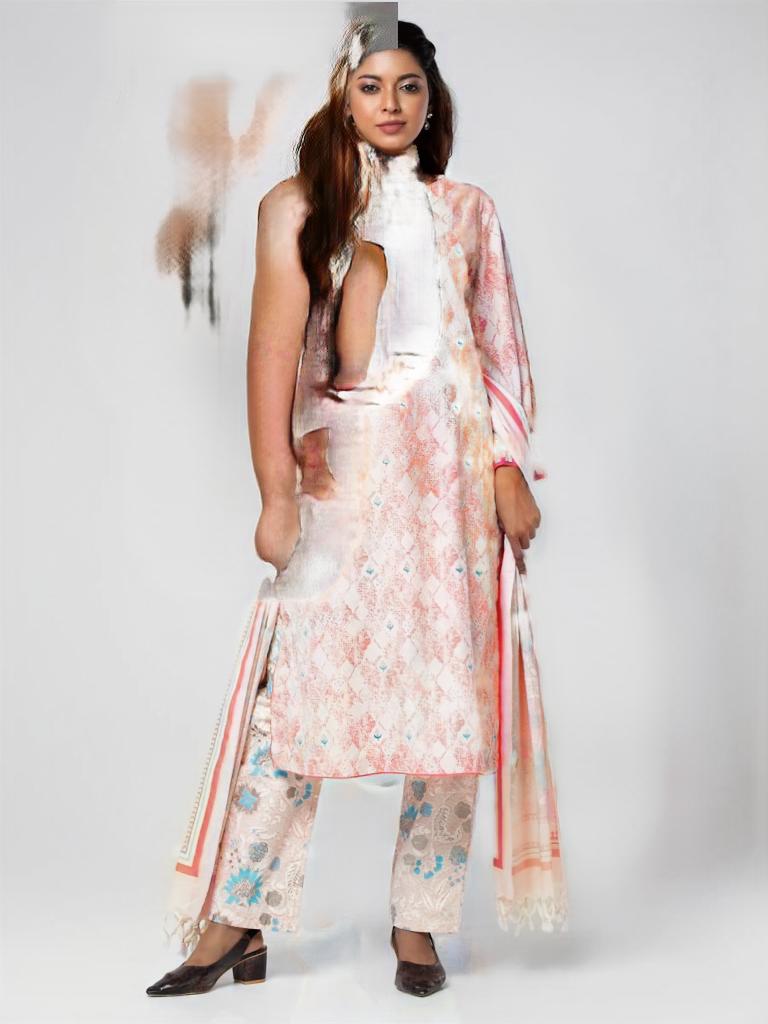} &
        \includegraphics[width=0.11\textwidth]{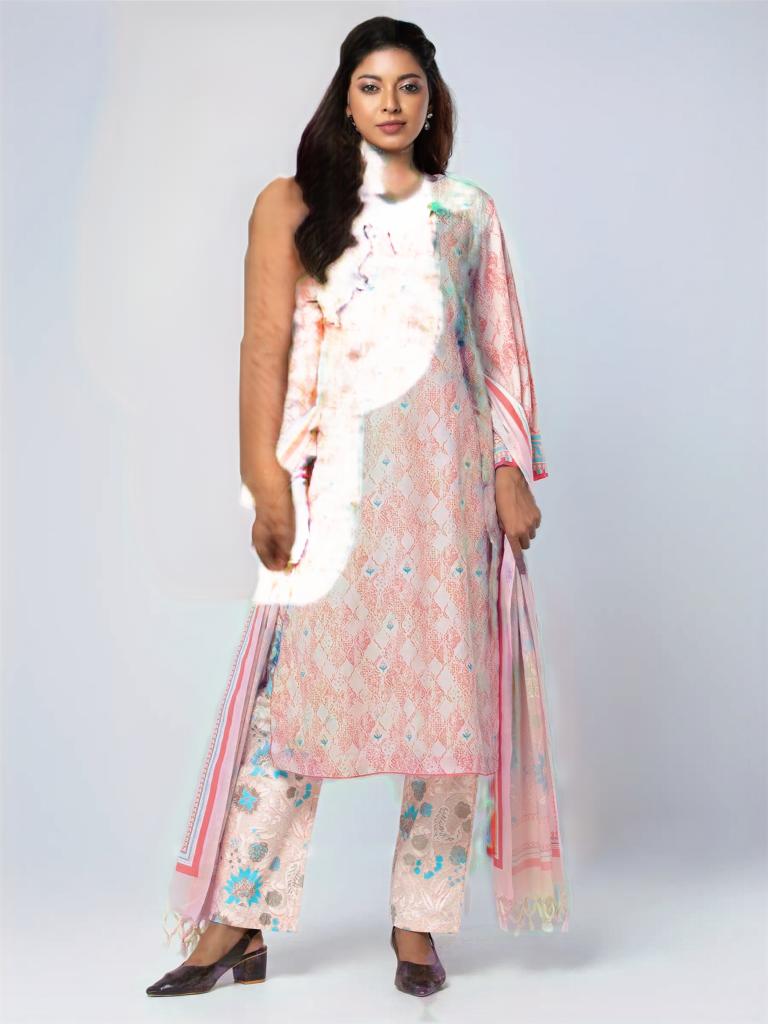} &
        \includegraphics[width=0.11\textwidth]{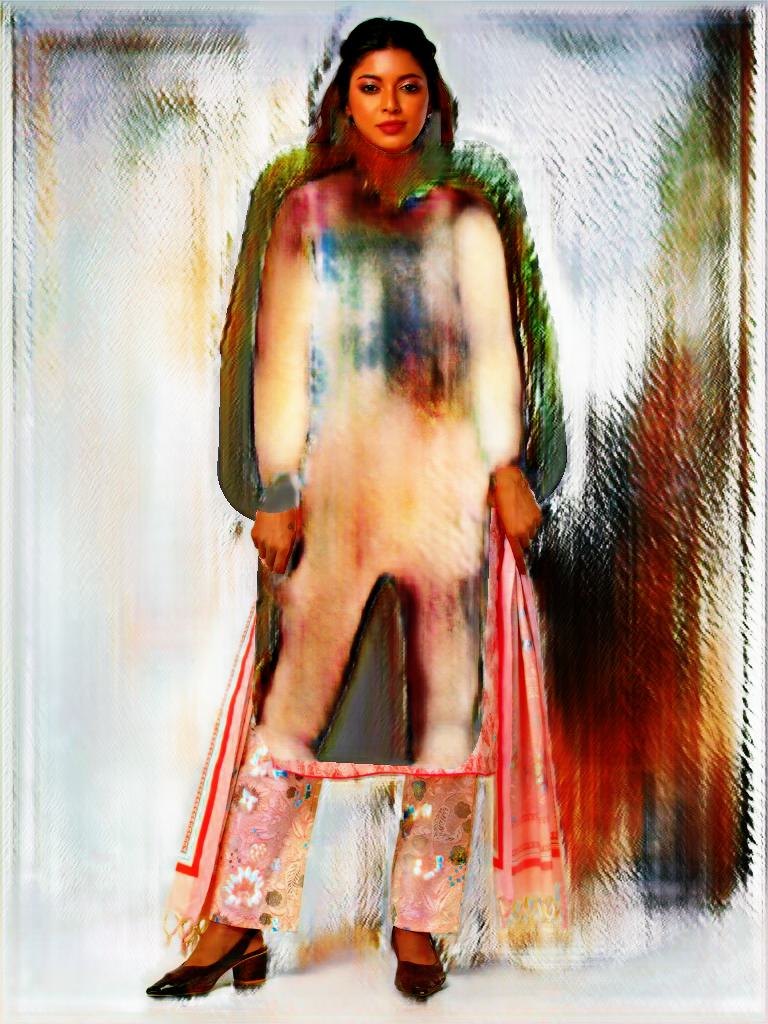} &
        \includegraphics[width=0.11\textwidth]{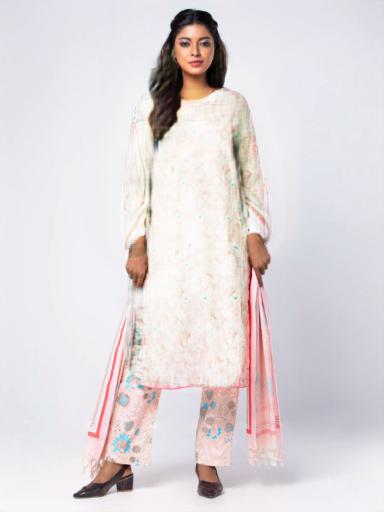} &
        \includegraphics[width=0.11\textwidth]{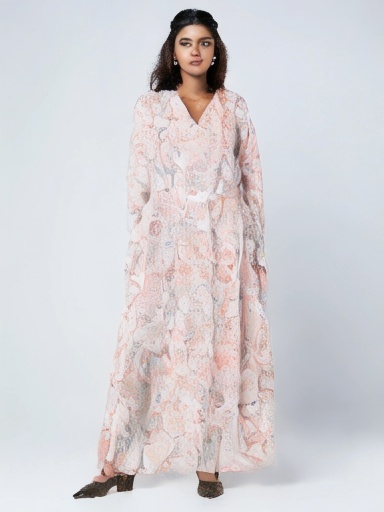} &
        \includegraphics[width=0.11\textwidth]{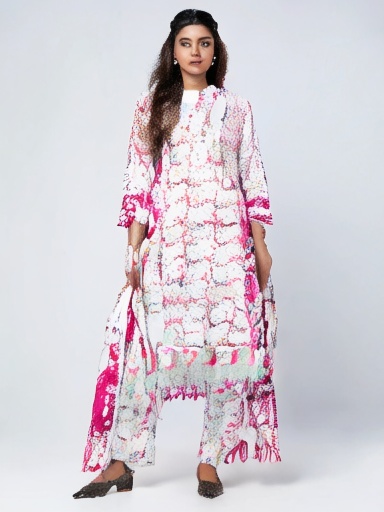} \\[4pt]

        \small Person & \small Target Cloth &
        \multicolumn{2}{c|}{\small VITON-HD} &
        \multicolumn{2}{c|}{\small HR-VITON} &
        \multicolumn{2}{c}{\small Stable VITON} \\
        & & \small Zero-Shot & \small Trained & \small Zero-Shot & \small Trained & \small Zero-Shot & \small Trained \\
    \end{tabular}
    \caption{Qualitative comparison of virtual try-on results across three garment types: saree (row 1), panjabi (row 2), and kameez (row 3). Each method shows zero-shot (left) and trained (right) results.}
    \label{fig:tryon_comparison}
\end{figure*}

The qualitative comparison highlights different behaviors among the three models on the custom dataset. StableVITON shows overfitting in the zero-shot setting where the garment structure is altered into a more western-style garment. After fine-tuning, the model produces more coherent results, though some color inconsistencies remain. HR-VITON struggles with garment deformation as the Try-On Condition Generator (TOCG) fails to properly learn cloth warping, resulting in noticeable misalignment with the body. In contrast, VITON-HD produces more stable outputs since it relies on a U-Net segmentation generator and TPS-based cloth warping, avoiding the severe warping failures seen in HR-VITON, although its results are generally less detailed than those from the diffusion-based StableVITON model.

\section{Future Work}
While BD-VITON attempts to address a significant research gap by introducing a different culture of clothing which inhibit more complex garment structure and intricate patterns, it still has room for improvements in certain aspects:
\begin{enumerate}
    \item \textbf{Limited garment diversity:} BD-VITON currently contains only three types of garments, which limits the diversity of the dataset. To address this limitation, we plan to expand the dataset by including culturally diverse garments such as the Japanese kimono and haori, Chinese hanfu, and other multi-layered clothing that exhibit similar structural complexity.
    \item \textbf{Limited dataset size:} BD-VITON has only 1,015 samples across both training and testing sets, which limits its scale. As a result, only fine-tuning of existing virtual try-on systems is feasible, while training generative models from scratch would likely lead to overfitting. To address this limitation, we plan to expand the dataset with additional samples.
\end{enumerate}

\section{Conclusion}
We believe that BD-VITON is a significant contribution to the field of try-on research. It introduces state-of-the-art architectures to clothing that have much more complex garment structures than Western garment based datasets. With adequate training, we are able to show that our dataset is effective in improving the performance of these models since they show much better results than zero shot inferences. We believe that with further refinement and proper training on BD-VITON, the generalization ability of try-on models can be further expanded to suit clothing of a wider diversity such that it performs optimally on any type of clothing.

\section*{Acknowledgements}
The dataset of BD-VITON has been collected by following the same policies as the original VITON dataset~\cite{han2018viton}. We are grateful that Aarong, Le Reve and Dorjibari have made their product images publicly accessible.

\bibliographystyle{IEEEtran}
\bibliography{main}

\end{document}